\pdfoutput=1
\documentclass{article}

\usepackage{microtype}
\usepackage{graphicx}
\usepackage{subcaption}
\usepackage{booktabs} % for professional tables
\usepackage{multirow} % used by tables/*.tex
\usepackage{xcolor}   % \color{gray} in tables/main_results.tex

\usepackage{hyperref}
\usepackage{enumitem}

\usepackage[preprint]{icml2027}

\usepackage{amsmath}
\usepackage{amssymb}
\usepackage{mathtools}
\usepackage{amsthm}
\usepackage{bm}

\usepackage[capitalize,noabbrev]{cleveref}

\theoremstyle{plain}
\newtheorem{theorem}{Theorem}[section]
\newtheorem{proposition}[theorem]{Proposition}

\theoremstyle{definition}

\theoremstyle{remark}

\usepackage[disable,textsize=tiny]{todonotes}

\newcommand{\R}{\mathbb{R}}
\newcommand{\E}{\mathbb{E}}

\newcommand{\Bern}{\operatorname{Bern}}
\newcommand{\sigmoid}{\sigma}
\newcommand{\bc}{{\bm{c}}}
\newcommand{\bx}{{\bm{x}}}
\newcommand{\bb}{{\bm{b}}}
\newcommand{\bmu}{{\bm{\mu}}}

\newcommand{\cS}{{\mathcal{S}}}
\newcommand{\Sbar}{{\bar{\mathcal{S}}}}
\newcommand{\pdrop}{{p_{\mathrm{drop}}}}

\newcommand{\CubLCAccFive}{96.9}

\newcommand{\CubSCBMAccFive}{87.5}

\newcommand{\CubCBMAccFive}{76.8}

\icmltitlerunning{Latent Class Concept Bottleneck Models}

\begin{document}

\twocolumn[
  \icmltitle{Latent Class Concept Bottleneck Models}



  \icmlsetsymbol{equal}{*}

  \begin{icmlauthorlist}
    \icmlauthor{Seokhun Park}{*}
  \end{icmlauthorlist}

  \icmlaffiliation{*}{Department of Statistics, Seoul National University, Seoul, Re
public of Korea}

  \icmlcorrespondingauthor{Seokhun Park}{shrdid@snu.ac.kr}

  \icmlkeywords{Machine Learning, ICML, Concept Bottleneck Models, Latent Class Models, Interpretability}

  \vskip 0.3in
]



\printAffiliationsAndNotice{}  % no special notice (required even if empty)

\begin{abstract}
Concept bottleneck models (CBMs) are interpretable image models that predict a target through a layer of human-understandable concepts.
Because the prediction depends on
the input only through these concepts, a user can correct mispredicted concepts
at test time and thereby change the prediction. Standard CBMs, however, treat
concepts as independent, so a correction never propagates to other concepts.
Recently, Stochastic CBMs (SCBMs) have modeled concept dependencies with a
Gaussian on the concept logits, but their interventions rely on a heuristic
rather than exact conditioning.
We propose \emph{Latent Class CBMs} (LC-CBMs), which model the concepts as an input-dependent mixture of product-Bernoulli distributions.
This gives an exact likelihood and exact Bayesian interventions.
%costing $O(K)$ per revealed concept. 
Across synthetic data, CUB, CIFAR-10 and CIFAR-100, LC-CBM matches or outperforms the baselines in target and concept accuracy, and achieves the most effective interventions on all four datasets.
We further show that the latent-class structure is not specific to CBMs: added to concept embedding models, it also improves their interventions.
\end{abstract}

% =============================================================================================
\section{Introduction}
\label{sec:intro}

Concept bottleneck models (CBMs, \citet{koh2020concept}) are interpretable models,
most commonly used for image classification, that explain their predictions through human-understandable concepts, such as `wing color' or `beak shape of
a bird'.
Formally, a CBM first maps an input $\bx$ to a vector of $C$ binary concepts
$\bc\in\{0,1\}^C$ and then predicts the target $y$ from $\bc$ alone.
Because the prediction passes through the concepts, a user can inspect them and
\emph{intervene}, replacing a mispredicted concept by its true value.

Since the original CBM, many extensions have been proposed to remove the need for concept annotations \cite{oikarinen2023labelfree,rao2024discover,srivastava2024vlg}, to close the accuracy gap to black-box models \cite{shang2024incremental,yuksekgonul2022post} and to improve interventions \cite{vandenhirtz2024stochastic, xu2024energy}.
Most of these models, however, still treat concepts as independent given the input, so an intervention changes only the corrected concept.
Real concepts are strongly dependent: in bird images, a black crown usually comes with a black throat and black underparts, so learning that one is present should change the prediction for the others.

Some works have tried to solve this problem by modeling the dependencies between concepts.
Autoregressive CBMs \citep{havasi2022addressing} factorize $p(\bc\mid\bx)$ in a fixed order of the concepts.
This gives an exact likelihood, but it has two limitations.
(1) Training and inference are sequential in $C$.
(2) Interventions reveal concepts in an arbitrary order, not in the order of the model, and conditioning on such concepts is intractable and must be approximated.
Stochastic CBMs \citep{vandenhirtz2024stochastic} place a multivariate Gaussian on the concept logits.
This makes the dependence structure explicit, but it has three limitations.
(1) The concept likelihood has no closed form, and its Monte Carlo estimate is biased.
(2) An intervened concept is replaced by a single logit value found by
optimization, and the other concepts are updated from this value rather than
from the revealed concept itself.
(3) SCBM predicts a full covariance matrix for each input, so the network must output $O(C^2)$ values per input.
This makes the model large and its interventions expensive, especially when $C$ is large (See \cref{app:cost}).

We address this with a discrete latent variable that explains the dependencies
between concepts, using the latent class model
\cite{goodman1974exploratory}.
The concepts are conditionally independent given the latent class, yet marginally dependent, and with enough classes such a model can represent any joint distribution of binary concepts \cite{dunson2009nonparametric}.
We call the resulting models \emph{Latent Class Concept Bottleneck Models} (LC-CBMs). 

Our contributions are:

\begin{itemize}
\item \textbf{Model.} We propose LC-CBMs, which model the dependencies between
concepts with a discrete latent class: the concepts are independent within
each latent class but dependent overall. 
LC-CBMs have an exact likelihood, and correcting a concept updates the other concepts exactly, with a closed-form update that costs $O(K)$ per revealed concept.

\item \textbf{No leakage.}
LC-CBMs use binary concepts, as in the hard CBM, and we show empirically that
they do not suffer from concept leakage.

\item \textbf{Generality.} The latent-class structure can also be added to other CBMs, such as CEM,
where it improves interventions.

\item \textbf{Empirical results.} On synthetic data, CUB, CIFAR-10 and CIFAR-100, LC-CBMs give the most effective interventions on every dataset and match or outperform the baselines in target and concept accuracy.
\end{itemize}

% =============================================================================================
\section{Background}
\label{sec:background}

\subsection{Concept Bottleneck Models}
Concept Bottleneck Models (CBMs; \citealp{koh2020concept}) predict a target
$y\in\{1,\dots,Y\}$ from an input image $\bx$ in two stages.
They first predict a set of $C$ predefined, human-interpretable concepts
$\bc\in\{0,1\}^C$ from the image.
Specifically, a neural-network encoder maps the image to a representation
$h(\bx)\in\R^d$, and a linear layer maps this representation to one logit per
concept,
\begin{equation*}
\bmu(\bx)=W h(\bx)+\mathbf b\in\R^C,\qquad W\in\R^{C\times d},\ \mathbf b\in\R^C,
\end{equation*}
where $\mu_j(\bx)$, the $j$-th entry of $\bmu(\bx)$, is the logit of concept
$j$.
The standard CBM treats the concepts as conditionally independent given the
image,
\begin{equation}
p_\theta(\bc\mid\bx)=\prod_{j=1}^C\Bern\bigl(c_j;\sigmoid(\mu_j(\bx))\bigr),
\label{eq:hard-cbm}
\end{equation}
where $\sigmoid(\cdot)$ is the sigmoid function and $\theta$ collects the parameters of the encoder and of the linear layer.
A head $p_\phi(y\mid\bc)$ with parameters $\phi$ then predicts the target from
these concepts alone.
Given training triples $(\bx,\bc,y)$, the model is trained by minimizing
\begin{align}
\mathcal{L}
&=-\log\sum_{\bc'\in\{0,1\}^C}p_\phi(y\mid\bc')\,p_\theta(\bc'\mid\bx) \nonumber\\
&\quad  +\lambda\sum_{j=1}^C\bigl(-\log p_\theta(c_j\mid\bx)\bigr),
\label{eq:cbm-loss}
\end{align}
where the first term is the target loss, obtained by marginalizing over the predicted concepts, the second is the concept loss, and $\lambda>0$ controls the relative weight of the concept loss.

CBMs differ in what the head receives as input.
A \emph{hard} CBM passes binary concept vectors $\bc'\in\{0,1\}^C$ to the
head, as in \eqref{eq:cbm-loss}, so that the prediction depends on the image
only through the presence or absence of each concept.
A \emph{soft} CBM instead passes the predicted probabilities
$\sigmoid(\bmu(\bx))$ directly to the head and is trained with the target loss
$-\log p_\phi\bigl(y\mid\sigmoid(\bmu(\bx))\bigr)$.

Because the final prediction depends on the image only through the predicted concepts, a user can interpret the decision process and \emph{intervene} at test time by replacing a mispredicted concept with its true value.
When the true values $\bc_\cS$ of a subset $\cS\subseteq\{1,\dots,C\}$ are intervened, the model clamps them in the head's input and draws the remaining concepts from $p_\theta(\bc_\Sbar\mid\bx,\bc_\cS)$.
Despite these advantages, CBMs have three main limitations, each of which has motivated a line of work.

\paragraph{Concept annotations.}
First, training a CBM requires concept labels for every image, which are costly to obtain.
Label-free and post-hoc CBMs remove this requirement \citep{oikarinen2023labelfree, yuksekgonul2022post}.
LF-CBM \citep{oikarinen2023labelfree}, for example, generates a concept set with a language model, filters it, and learns a concept layer on a frozen backbone by matching CLIP image--concept similarities \citep{radford2021learning}; a sparse linear head is then fitted on the standardized continuous concept activations.
Its concepts are therefore never binary labels, and an intervention has to be defined by replacing an activation.
Moreover, the resulting concept supervision is only as reliable as CLIP, which may not reflect the concepts actually present in an individual image \citep{yun2022vision, srivastava2024vlg}.

\paragraph{Concept leakage.}
Second, when a Soft CBM is trained end-to-end, the predicted concepts can encode information about the target beyond the presence or absence of each concept \citep{margeloiu2021concept, mahinpei2021promises}.
The head then relies on this leaked information rather than on the concepts themselves, which undermines both the interpretation of the concepts and the effect of interventions.
Proposed remedies include training the concept predictor and the head
separately, and using binary instead of soft concepts, i.e., Hard CBM.
Hard CBMs, however, often lose target accuracy, since binary concepts discard the information that soft probabilities carry.

\paragraph{Concept dependencies.}
Finally, CBMs treat the concepts as conditionally independent given the image, so that $p_\theta(\bc_\Sbar\mid\bx,\bc_\cS)=p_\theta(\bc_\Sbar\mid\bx)$: an intervention changes only the revealed concepts and carries no information about the others.
Several works model these dependencies.
Autoregressive CBMs \citep{havasi2022addressing} model the dependencies by
factorizing $p_\theta(\bc\mid\bx)$ sequentially. 
This makes training and
inference sequential in $C$, and conditioning on a revealed concept updates only the concepts that come after it in the chosen order; updating the earlier ones requires sampling.
Stochastic CBMs (SCBMs; \citealp{vandenhirtz2024stochastic}) instead replace the deterministic logits $\bmu(\bx)=(\mu_1(\bx),\dots,\mu_C(\bx))^\top$ by
random logits $\bm\eta=(\eta_1,\dots,\eta_C)^\top$ with a multivariate Gaussian
distribution,
\begin{align}
\begin{split}
&\bm\eta\mid\bx\sim\mathcal N\bigl(\bmu(\bx),\Sigma(\bx)\bigr), \\
&c_j\mid\bm\eta\sim\Bern\bigl(\sigmoid(\eta_j)\bigr),
\end{split}
\label{eq:scbm}
\end{align}
independently, so that the concepts become dependent through $\Sigma(\bx)$.
The covariance is either a global parameter or predicted from $h(\bx)$ through its Cholesky factor (amortized).
For an intervention, SCBM first finds the logits of the revealed concepts that best explain their values within a 99\% confidence region,
\begin{align*}
\hat{\bm\eta}_\cS=\arg\max_{\bm\eta_\cS}\ &\log p(\bc_\cS\mid\bm\eta_\cS)\\
\text{s.t.}\ &(\bm\eta_\cS-\bmu_\cS)^\top\Sigma_{\cS\cS}^{-1}(\bm\eta_\cS-\bmu_\cS)\le\chi^2_{|\cS|}(0.99),
\end{align*}
and then samples the remaining logits from the Gaussian conditional $\bm\eta_\Sbar\mid\bm\eta_\cS=\hat{\bm\eta}_\cS$.
The other concepts are thus updated from a point estimate, not from the revealed concepts themselves.
This shortcut, like the Monte Carlo estimate of the likelihood, is needed because the Gaussian integral over $\bm\eta$ has no closed form.
As a result, SCBM trains on a biased likelihood, its interventions are not
exact, and each intervention requires an optimization per sample, which makes
interventions computationally expensive (See Appendix \ref{app:cost}).
Energy-based CBMs \citep{xu2024energy} model dependencies with a joint energy over images, concepts and targets, but they also lack an exact likelihood, and each intervention requires an optimization.
A comparison with Energy-based CBMs is given in Appendix \ref{app:ecbm}.

\begin{figure*}[t]
    \centering
    \includegraphics[width=0.95\linewidth]{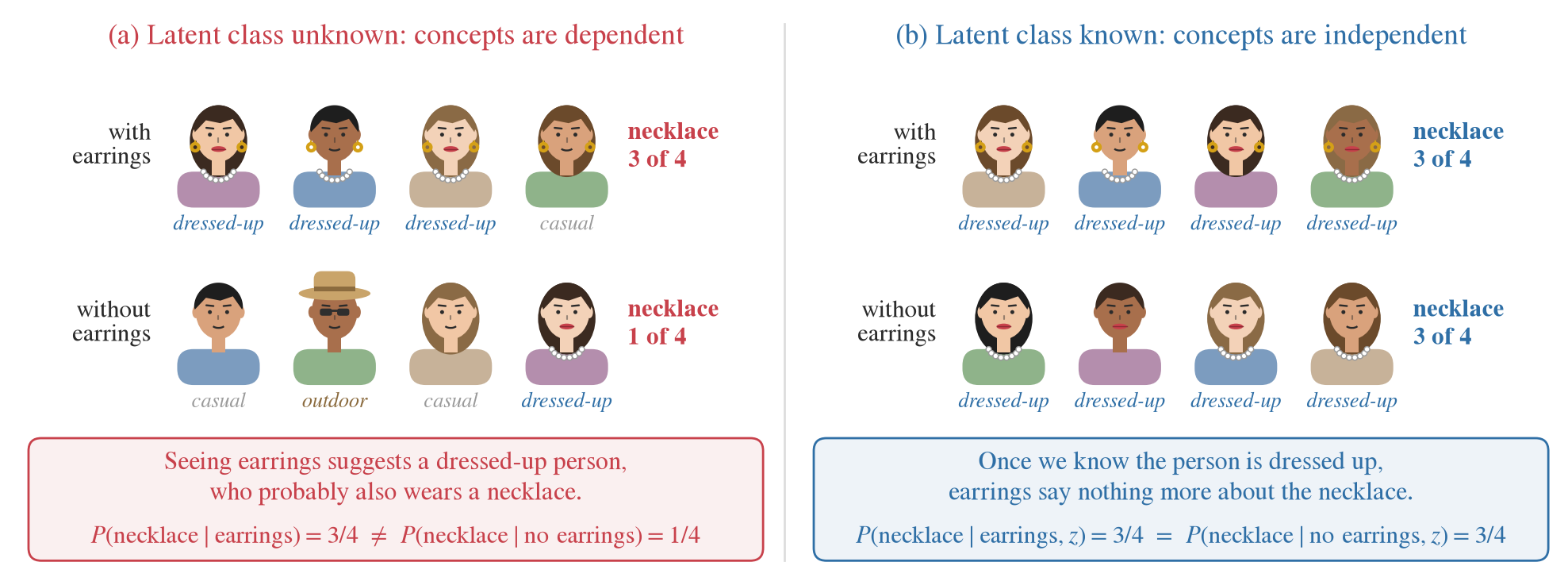}
    \caption{Local independence on face attributes. The italic label under each face is its latent class.
    (a) Latent class unknown: earrings and a necklace are dependent.
    (b) Latent class known: they are independent.}
    \label{fig:latent_class}
\end{figure*}

\section{Proposed Model}
\label{sec:model}

\subsection{Latent-Class Concept Bottleneck Models}
\label{sec:method}

To address the limitations of existing models discussed in \cref{sec:background}, we propose \emph{Latent Class Concept Bottleneck Models} (LC-CBMs).
Instead of a continuous Gaussian on the concept logits, LC-CBM explains the dependencies between concepts with a discrete latent variable $z\in\{1,\dots,K\}$, which we call the \emph{latent class} to distinguish it from the target class $y$.
Each latent class represents a typical combination of concepts, such as
``black wings and a black body'', which we call its \emph{concept profile}.
The latent classes are not supervised by $y$.

The key assumption is that the concepts are conditionally independent given the latent class.
This assumption is natural when concepts co-occur because of a common cause rather than because one concept determines another.
\cref{fig:latent_class} illustrates this with face attributes.
Earrings and a necklace tend to appear together, but wearing one does not cause wearing the other; both are typical of a \emph{dressed-up} appearance.
When the latent class is unknown (\cref{fig:latent_class}a), the two concepts are therefore dependent: three of four people with earrings wear a necklace, but only one of four without earrings does, because seeing earrings suggests a dressed-up person, who probably also wears a necklace.
Once the latent class is known (Figure \ref{fig:latent_class}b), each item is a separate choice of the person, and earrings say nothing more about the necklace: among dressed-up people, the proportion wearing a necklace is the same with and without earrings.
The dependence between concepts thus arises only from uncertainty about the latent class.
Note that a latent class describes an appearance rather than the target: a dressed-up appearance can occur with any target class.

LC-CBM builds this assumption directly into the concept distribution.
Given an image $\bx$, it first predicts which latent class the image belongs to,
\begin{equation}
\pi_k(\bx) := p(z=k\mid\bx) = \operatorname{softmax}_k\bigl(W_\pi h(\bx)+\bm a\bigr),
\label{eq:pi}
\end{equation}
where $W_\pi\in\R^{K\times d}$ and $\bm a\in\R^K$.
It then predicts every concept independently within each latent class,
\begin{equation}
p_{kj}(\bx) := p(c_j=1\mid z=k,\bx) = \sigmoid\bigl(\mu_j(\bx)+b_{kj}\bigr),
\label{eq:shift}
\end{equation}
where $\mu_j(\bx)$ is the $j$-th entry of $\bmu(\bx)=W_\mu h(\bx)+\bm\beta$, with $W_\mu\in\R^{C\times d}$ and $\bm\beta\in\R^C$, and gives the logit of concept $j$ from the image, and $\bb_k=(b_{k1},\dots,b_{kC})\in\R^C$ is a class-specific bias vector that encodes the concept profile of latent class $k$.
Summing over the latent classes gives the concept distribution, a mixture of $K$ product-Bernoulli distributions:
\begin{equation}
p_\theta(\bc\mid\bx)=\sum_{k=1}^K\pi_k(\bx)\prod_{j=1}^C\Bern\bigl(c_j;p_{kj}(\bx)\bigr).
\label{eq:lc}
\end{equation}
Here, $\theta$ denotes all learnable parameters of the concept distribution, including the encoder $h$.
The concepts are independent within each latent class, but dependent once the latent classes are mixed.

Finally, as in CBM, the target is predicted by a head that is linear in the concepts, $p_\phi(y\mid\bc)=\operatorname{softmax}(W_y\bc+\bm w_0)$ with $W_y\in\R^{Y\times C}$ and $\bm w_0\in\R^Y$, so that the target depends on the image only through $\bc$.
Summing over the possible values $k$ of the latent class $z$, the prediction is
\begin{equation}
p(y\mid\bx) = \sum_{k=1}^K\pi_k(\bx)\,\E_{\bc\sim\prod_j\Bern(p_{kj}(\bx))}\bigl[p_\phi(y\mid\bc)\bigr],
\label{eq:predict}
\end{equation}
where the expectation is over the concept distribution of class $k$; the derivation is given in \cref{app:predict}.
The sum over the $K$ latent classes is computed exactly, so $z$ never needs to be sampled, and the expectation is estimated with $M$ concept samples per class (we use $M=10$).

\subsection{Exact interventions}
\begin{proposition}[Closure under conditioning]
\label{prop:closure}
Let the values $\bc_\cS$ of a subset $\cS\subseteq\{1,\dots,C\}$ of concepts be intervened, and let $\Sbar=\{1,\dots,C\}\setminus\cS$ denote the unintervened concepts.
For each latent class $k$, let
\begin{equation}
L_k(\bx,\bc_\cS)=\prod_{j\in\cS}p_{kj}(\bx)^{c_j}\bigl(1-p_{kj}(\bx)\bigr)^{1-c_j}
\label{eq:class-lik}
\end{equation}
be the likelihood of the revealed concepts under class $k$.
Then $p(\bc_\Sbar\mid\bx,\bc_\cS)$ is again a latent class model of the form \eqref{eq:lc} over $\bc_\Sbar$, with the same class-conditional probabilities $p_{kj}(\bx)$, $j\in\Sbar$, and posterior class weights
\begin{equation}
\pi_k(\bx,\bc_\cS):=p(z=k\mid\bx,\bc_\cS)
=\frac{\pi_k(\bx)\,L_k(\bx,\bc_\cS)}{\sum_{l=1}^K\pi_l(\bx)\,L_l(\bx,\bc_\cS)}.
\label{eq:posterior}
\end{equation}
In particular, for every $j\in\Sbar$,
\begin{equation}
p(c_j=1\mid\bx,\bc_\cS)=\sum_{k=1}^K\pi_k(\bx,\bc_\cS)\,p_{kj}(\bx).
\label{eq:post-marginal}
\end{equation}
\end{proposition}

In words, intervening on concepts does not change the class-conditional probabilities $p_{kj}(\bx)$; it only reweights the latent classes.
Latent classes that agree with the intervened concepts become more likely, and the remaining concepts are predicted from these updated classes.
In \cref{fig:latent_class}, for example, intervening on earrings moves weight toward the dressed-up class and thereby raises the probability of a necklace.

This update is exact and has no tuning parameter.
It is also computationally efficient, which matters because a user intervenes on concepts one at a time and the model must update after each intervention.
Each time a concept is intervened on, $L_k$ is multiplied by that concept's probability under class $k$, so updating the class posterior requires only one multiplication per class, that is, $O(K)$ operations.
The marginals of all remaining concepts then follow from \eqref{eq:post-marginal} in $O(KC)$ operations, with no optimization or sampling.

In contrast, SCBM can perform this update neither exactly nor efficiently.
For SCBM, the exact update would be
\begin{equation}
p(\bc_\Sbar\mid\bx,\bc_\cS)
=\int\prod_{j\in\Sbar}\Bern\bigl(c_j;\sigmoid(\eta_j)\bigr)\,
p(\bm\eta\mid\bx,\bc_\cS)\,d\bm\eta,
\label{eq:scbm-cond}
\end{equation}
where the posterior of the logits is
\begin{equation}
p(\bm\eta\mid\bx,\bc_\cS)\;\propto\;\mathcal N\bigl(\bm\eta;\bmu(\bx),\Sigma(\bx)\bigr)\prod_{j\in\cS}\Bern\bigl(c_j;\sigmoid(\eta_j)\bigr).
\label{eq:scbm-post}
\end{equation}
This posterior is not Gaussian, and its normalizing constant is an integral over $\bm\eta$ with no closed form, so neither equation can be computed exactly.
SCBM therefore uses a heuristic: for every sample at every intervention step, it solves a constrained optimization problem to pick a single logit value $\hat{\bm\eta}_\cS$ for the revealed concepts within a 99\% confidence region, and then samples the remaining logits from the Gaussian conditioned on $\hat{\bm\eta}_\cS$.
The result is an approximation that conditions on $\hat{\bm\eta}_\cS$ rather than on the revealed concepts, and it is far more expensive than the closed-form update of LC-CBM (See Appendix \ref{app:cost}).

After an intervention, the target is predicted with (\ref{eq:predict}), using the updated weights $\pi_k(\bx,\bc_\cS)$ in place of $\pi_k(\bx)$ and setting the revealed concepts to their given values.

\begin{table*}[t]
\centering
\caption{Prediction performance (mean$\pm$std over seeds, \%). Gray rows ($^\ast$) are the numbers reported by \citet{vandenhirtz2024stochastic} (Tables~1 and~3). Concept Brier and ECE follow the SCBM definition (per concept, averaged). Bold: best among our runs.}
\label{tab:main}
\resizebox{0.9\textwidth}{!}{\begin{tabular}{llccccc}
\toprule
Dataset & Method & Concept acc. $\uparrow$ & Target acc. $\uparrow$ & Concept Brier $\downarrow$ & Concept ECE $\downarrow$ & Joint NLL $\downarrow$ \\
\midrule
\multirow{6}{*}{Synthetic} & Soft CBM & 61.96{\scriptsize$\pm$1.56} & 58.25{\scriptsize$\pm$6.91} & 28.40{\scriptsize$\pm$1.04} & 21.98{\scriptsize$\pm$0.96} & 97.40{\scriptsize$\pm$2.77} \\
 & Hard CBM & 61.45{\scriptsize$\pm$1.97} & 58.78{\scriptsize$\pm$5.63} & 28.76{\scriptsize$\pm$1.35} & 22.33{\scriptsize$\pm$1.29} & 98.50{\scriptsize$\pm$3.78} \\
 & {\color{gray}Autoregressive CBM$^\ast$} & {\color{gray}62.17} & {\color{gray}59.60} & {\color{gray}24.84} & {\color{gray}13.54} & {\color{gray}---} \\
 & Global SCBM & 62.03{\scriptsize$\pm$1.53} & 58.08{\scriptsize$\pm$6.90} & 27.34{\scriptsize$\pm$1.04} & 19.67{\scriptsize$\pm$1.01} & 65.87{\scriptsize$\pm$2.26} \\
 & Amortized SCBM & 62.41{\scriptsize$\pm$2.08} & 59.12{\scriptsize$\pm$5.82} & 25.55{\scriptsize$\pm$1.33} & 15.49{\scriptsize$\pm$1.61} & 65.05{\scriptsize$\pm$3.10} \\
 & \textbf{LC-CBM (ours)} & \textbf{62.67{\scriptsize$\pm$1.92}} & \textbf{59.23{\scriptsize$\pm$5.96}} & \textbf{24.15{\scriptsize$\pm$0.95}} & \textbf{11.78{\scriptsize$\pm$0.85}} & \textbf{52.98{\scriptsize$\pm$1.94}} \\
\midrule
\multirow{6}{*}{CUB} & Soft CBM & 95.20{\scriptsize$\pm$0.07} & 67.54{\scriptsize$\pm$0.45} & 3.77{\scriptsize$\pm$0.07} & 2.38{\scriptsize$\pm$0.08} & 16.46{\scriptsize$\pm$0.38} \\
 & Hard CBM & 95.15{\scriptsize$\pm$0.09} & 67.26{\scriptsize$\pm$0.78} & 3.80{\scriptsize$\pm$0.07} & 2.42{\scriptsize$\pm$0.09} & 16.63{\scriptsize$\pm$0.36} \\
 & {\color{gray}Autoregressive CBM$^\ast$} & {\color{gray}95.33} & {\color{gray}69.24} & {\color{gray}3.75} & {\color{gray}2.73} & {\color{gray}---} \\
 & Global SCBM & 95.15{\scriptsize$\pm$0.06} & 67.72{\scriptsize$\pm$0.27} & 3.75{\scriptsize$\pm$0.04} & 2.31{\scriptsize$\pm$0.05} & 12.96{\scriptsize$\pm$0.14} \\
 & Amortized SCBM & 95.44{\scriptsize$\pm$0.10} & 69.85{\scriptsize$\pm$0.69} & 3.49{\scriptsize$\pm$0.08} & 1.79{\scriptsize$\pm$0.10} & 9.17{\scriptsize$\pm$0.23} \\
 & \textbf{LC-CBM (ours)} & \textbf{96.07{\scriptsize$\pm$0.05}} & \textbf{72.35{\scriptsize$\pm$0.40}} & \textbf{3.01{\scriptsize$\pm$0.03}} & \textbf{1.09{\scriptsize$\pm$0.04}} & \textbf{1.21{\scriptsize$\pm$0.01}} \\
\midrule
\multirow{6}{*}{CIFAR-10} & Soft CBM & 85.42{\scriptsize$\pm$0.06} & 69.60{\scriptsize$\pm$0.26} & 10.49{\scriptsize$\pm$0.06} & 5.02{\scriptsize$\pm$0.17} & 49.11{\scriptsize$\pm$0.42} \\
 & Hard CBM & 85.41{\scriptsize$\pm$0.06} & 68.82{\scriptsize$\pm$0.34} & 10.52{\scriptsize$\pm$0.06} & 5.17{\scriptsize$\pm$0.19} & 49.42{\scriptsize$\pm$0.43} \\
 & {\color{gray}Autoregressive CBM$^\ast$} & {\color{gray}85.31} & {\color{gray}68.88} & {\color{gray}10.70} & {\color{gray}6.07} & {\color{gray}---} \\
 & Global SCBM & 85.80{\scriptsize$\pm$0.05} & 70.07{\scriptsize$\pm$0.65} & 10.01{\scriptsize$\pm$0.05} & 3.00{\scriptsize$\pm$0.15} & 37.21{\scriptsize$\pm$0.34} \\
 & Amortized SCBM & 85.92{\scriptsize$\pm$0.04} & 70.89{\scriptsize$\pm$0.30} & 9.91{\scriptsize$\pm$0.02} & 2.35{\scriptsize$\pm$0.11} & 37.33{\scriptsize$\pm$0.35} \\
 & \textbf{LC-CBM (ours)} & \textbf{86.47{\scriptsize$\pm$0.02}} & \textbf{71.47{\scriptsize$\pm$0.29}} & \textbf{9.49{\scriptsize$\pm$0.02}} & \textbf{0.97{\scriptsize$\pm$0.07}} & \textbf{33.22{\scriptsize$\pm$0.03}} \\
\midrule
\multirow{5}{*}{CIFAR-100} & Soft CBM & 84.44{\scriptsize$\pm$0.02} & 39.33{\scriptsize$\pm$0.20} & 11.06{\scriptsize$\pm$0.03} & 4.16{\scriptsize$\pm$0.25} & 316.34{\scriptsize$\pm$2.12} \\
 & Hard CBM & 84.39{\scriptsize$\pm$0.03} & 38.02{\scriptsize$\pm$0.48} & 11.09{\scriptsize$\pm$0.01} & 4.16{\scriptsize$\pm$0.06} & 316.84{\scriptsize$\pm$0.16} \\
 & Global SCBM & 84.76{\scriptsize$\pm$0.04} & 39.84{\scriptsize$\pm$0.14} & 10.68{\scriptsize$\pm$0.03} & 2.48{\scriptsize$\pm$0.05} & 244.69{\scriptsize$\pm$0.90} \\
 & Amortized SCBM & 84.85{\scriptsize$\pm$0.04} & 38.19{\scriptsize$\pm$0.58} & 10.61{\scriptsize$\pm$0.03} & 1.67{\scriptsize$\pm$0.17} & 240.06{\scriptsize$\pm$0.06} \\
 & \textbf{LC-CBM (ours)} & \textbf{85.02{\scriptsize$\pm$0.03}} & \textbf{40.03{\scriptsize$\pm$0.47}} & \textbf{10.49{\scriptsize$\pm$0.02}} & \textbf{1.47{\scriptsize$\pm$0.09}} & \textbf{224.70{\scriptsize$\pm$0.25}} \\
\bottomrule
\end{tabular}
}
\end{table*}

\subsection{Training}
\label{sec:training}

Given a training data $(\bx,\bc,y)$, we minimize
\begin{equation}
\begin{aligned}
\mathcal L={}&-\log\hat p(y\mid\bx)+\lambda\bigl(-\log p_\theta(\bc\mid\bx)\bigr)\\
&+\alpha\bigl(-\log p_\phi(y\mid\bc)\bigr),
\end{aligned}
\label{eq:loss}
\end{equation}
which consists of a target loss, a concept loss and an auxiliary head loss.
The \emph{target loss} uses the prediction in (\ref{eq:predict}), with the inner expectation estimated from $M$ samples per latent class,
\begin{equation}
\hat p(y\mid\bx)=\sum_{k=1}^K\pi_k(\bx)\,\frac1M\sum_{m=1}^M p_\phi\bigl(y\mid\tilde\bc^{(k,m)}\bigr),
\label{eq:pred-mc}
\end{equation}
where $\tilde\bc^{(k,m)}$ are concept vectors sampled from class $k$.
To backpropagate through these binary samples, we use straight-through relaxed Bernoulli samples \citep{jang2016categorical,bengio2013estimating} with a temperature annealed from 1 to 0.5, as in \citet{vandenhirtz2024stochastic}.

The \emph{concept loss} is the exact negative log-likelihood of \eqref{eq:lc}, computed with a log-sum-exp over the $K$ latent classes.
If some concept labels are missing, we simply drop their factors from the product in \eqref{eq:lc}, which gives the exact negative log-likelihood $-\log p_\theta(\bc_{\mathrm{obs}}\mid\bx)$ of the observed labels.
Experiments with missing concept labels are reported in Appendix \ref{sec:robustness} and Appendix \ref{sec:lf}.
The \emph{auxiliary head loss} trains the head on the true concepts $\bc$, not only on estimated ones.
This matters because, once a user has corrected all concepts, the head receives the true concept vector as input.
Results with and without the auxiliary head loss are reported in Appendix \ref{app:auxhead}.
%We use $\lambda=1$ and $\alpha=1$ unless stated otherwise.

%=============================================================================================

% =============================================================================================

\section{Experiments}

\subsection{Setup}

\paragraph{Datasets.}
We evaluate on four datasets: Synthetic, CUB, CIFAR-10 and CIFAR-100.
\emph{Synthetic} data with $C=100$ binary concepts are generated as in \cite{vandenhirtz2024stochastic}.
\emph{CUB} \citep{wah2011cub} has 112 bird concepts, preprocessed as in \citet{koh2020concept}.
For \emph{CIFAR-10} and \emph{CIFAR-100} \citep{krizhevsky2009learning}, we use the concept sets of \citet{oikarinen2023labelfree} (143 and 892 concepts), labeled with CLIP \citep{radford2021learning}.
Further details are given in \cref{app:data}.

\paragraph{Baseline models.}
We compare LC-CBM with the Soft and Hard CBM \citep{koh2020concept}, CEM \citep{espinosa2022concept}, autoregressive CBM \citep{havasi2022addressing}, and two variants of SCBM \citep{vandenhirtz2024stochastic}: \emph{Global SCBM}, which learns a single covariance matrix shared by all inputs, and \emph{Amortized SCBM}, which predicts a covariance matrix for each input.

\paragraph{Metrics.}
All metrics are computed on the test set.
Let $\hat p_j(\bx)=p(c_j=1\mid\bx)$ be the predicted marginal probability of concept $j$.
\begin{itemize}[leftmargin=*,nosep]
\item \emph{Target accuracy}: the fraction of test images whose target is predicted correctly.
\item \emph{Concept accuracy}: the fraction of correctly predicted concepts, where concept $j$ is predicted as present if $\hat p_j(\bx)>0.5$.
\item \emph{Concept Brier score}: the mean squared error $(\hat p_j(\bx)-c_j)^2$; lower is better.
\item \emph{Concept ECE}: the expected calibration error \citep{naeini2015obtaining} of each concept, treated as a binary classification problem; lower is better.
\end{itemize}
Concept Brier score and ECE are computed per concept and then averaged, as in \citet{vandenhirtz2024stochastic}.

\begin{figure*}[t]
\centering
\includegraphics[width=\linewidth]{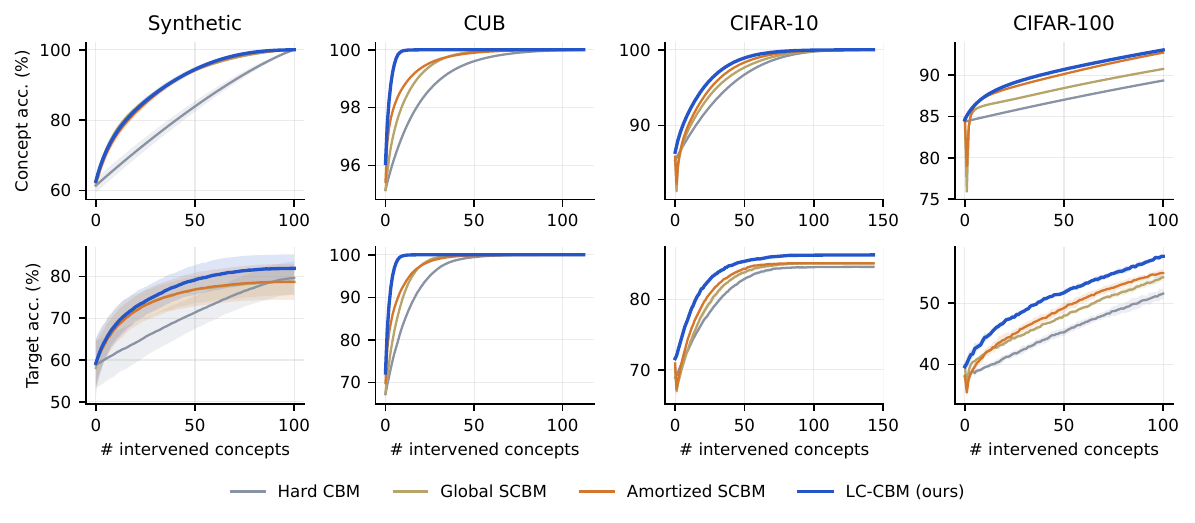}
\caption{Concept accuracy (top) and target accuracy (bottom) after intervening on concepts in order of predicted uncertainty. Lines: mean over seeds; bands: $\pm1$ std.}
\label{fig:interventions}
\end{figure*}

\subsection{Prediction Performance}
\label{sec:predict}
All models share the same encoder for a given dataset, trained end-to-end with the concept and target layers; the encoders and data augmentation follow the setting in \cite{vandenhirtz2024stochastic}.
That is, the backbone is a three-layer MLP (width 256) for the synthetic data, an ImageNet-pretrained ResNet-18 for CUB, and a small two-layer CNN for CIFAR-10 and CIFAR-100.

Table~\ref{tab:main} compares LC-CBM with Soft CBM, Hard CBM, CEM, Autoregressive CBM, Global SCBM and Amortized SCBM, averaged over five runs on CIFAR-100 and ten runs elsewhere.
Autoregressive CBM results are taken from \citet{vandenhirtz2024stochastic}; for the models we reran, our results are close to those reported there.
LC-CBM performs best on most datasets and metrics.
Details are in Appendix \ref{app:details}.

\subsection{Intervention performance}
\label{sec:interven}

\begin{table}[h]
\small
\centering
\caption{Target accuracy (\%) after revealing $n$ concepts in order of uncertainty, and the normalized area under the curve (AUC, mean$\pm$std over seeds).}
\label{tab:interventions}
\resizebox{0.5\textwidth}{!}{\begin{tabular}{lcccc}
\toprule
Method & Synthetic & CUB & CIFAR-10 & CIFAR-100 \\
\midrule
Hard CBM$^\dagger$       & 70.75{\scriptsize$\pm$4.04} & 95.98{\scriptsize$\pm$0.09} & 81.91{\scriptsize$\pm$0.09} & 45.15{\scriptsize$\pm$0.58} \\
Global SCBM$^\dagger$    & 74.94{\scriptsize$\pm$3.22} & 97.55{\scriptsize$\pm$0.06} & 82.54{\scriptsize$\pm$0.16} & 47.58{\scriptsize$\pm$0.27} \\
Amortized SCBM$^\dagger$ & 74.82{\scriptsize$\pm$3.77} & 97.96{\scriptsize$\pm$0.15} & 82.97{\scriptsize$\pm$0.11} & 48.45{\scriptsize$\pm$0.63} \\
\textbf{LC-CBM (ours)}   & \textbf{76.90{\scriptsize$\pm$3.40}} & \textbf{99.31{\scriptsize$\pm$0.01}} & \textbf{84.45{\scriptsize$\pm$0.06}} & \textbf{51.03{\scriptsize$\pm$0.27}} \\
\bottomrule
\end{tabular}}
\end{table}

The main advantage of CBMs is that they allow interventions at test time.
We therefore compare the intervention performance of LC-CBM with that of Hard CBM, Global SCBM and Amortized SCBM.
We consider the uncertainty-based intervention policy of \citet{shin2023closer}.
That is, at each step, the policy selects, among the concepts not yet intervened on, the one whose predicted probability $p_\theta(c_j=1\mid\bx)$ is closest to
$0.5$, i.e.\ the most uncertain one, and the user sets it to its true value.

Figure~\ref{fig:interventions} shows the concept accuracy and the target accuracy on the synthetic, CUB, CIFAR-10 and CIFAR-100 datasets as the number of intervened concepts increases.
Table~\ref{tab:interventions} summarizes each target-accuracy curve by its normalized area under the curve (AUC), i.e.\ the area under the curve divided by the range of the horizontal axis, which equals the average target accuracy over all intervention steps.
All curves and AUCs are averaged over five seeds on CIFAR-100 and ten seeds on the other datasets.
A higher AUC means that interventions improve the prediction more and earlier.
LC-CBM attains the highest AUC on every dataset, indicating that its interventions are the most effective.

The gain is largest on CUB, where five interventions raise target accuracy to \CubLCAccFive\%, against \CubSCBMAccFive\% for SCBM and \CubCBMAccFive\% for Hard CBM.
On CIFAR-10 and CIFAR-100, the accuracy of SCBM drops after the first few interventions, whereas that of LC-CBM increases steadily.
This difference comes from how the two models update the other concepts.
LC-CBM uses exact Bayesian updating, so an intervened concept changes the others only as much as it is related to them.
SCBM instead pushes the logit of the intervened concept to a fixed value and moves all other logits along with it.
When the first intervened concepts are very uncertain, this push is large, and it can move other concepts in the wrong direction.

\subsection{LC as a plug-in concept layer}
\label{sec:plugin}

\begin{table*}[t]
\centering
\caption{CEM vs.\ LC-CEM (CEM with latent-class concept probabilities).
Mean $\pm$ std.\ over five seeds; better values in bold.}
\label{tab:cem}
\resizebox{0.7\textwidth}{!}{\begin{tabular}{llcccc}
\toprule
Dataset & Model & Concept acc.\ $\uparrow$ & Target acc.\ $\uparrow$ & Concept ECE $\downarrow$ & AUC (unc.) $\uparrow$ \\
\midrule
\multirow{2}{*}{CUB}
 & CEM & 95.38{\scriptsize$\pm$0.05} & 69.57{\scriptsize$\pm$0.24} & 3.13{\scriptsize$\pm$0.02} & 87.45{\scriptsize$\pm$1.37} \\
 & LC-CEM & \textbf{95.86}{\scriptsize$\pm$0.05} & \textbf{71.14}{\scriptsize$\pm$0.25} & \textbf{0.77}{\scriptsize$\pm$0.02} & \textbf{99.25}{\scriptsize$\pm$0.02} \\
\midrule
\multirow{2}{*}{CIFAR-10}
 & CEM & 84.92{\scriptsize$\pm$0.19} & 70.79{\scriptsize$\pm$0.50} & 7.51{\scriptsize$\pm$0.61} & 79.14{\scriptsize$\pm$0.84} \\
 & LC-CEM & \textbf{86.68}{\scriptsize$\pm$0.03} & \textbf{74.39}{\scriptsize$\pm$0.32} & \textbf{1.36}{\scriptsize$\pm$0.27} & \textbf{87.21}{\scriptsize$\pm$0.32} \\
\midrule
\multirow{2}{*}{CIFAR-100}
 & CEM & 83.89{\scriptsize$\pm$0.09} & \textbf{42.36}{\scriptsize$\pm$0.61} & 6.96{\scriptsize$\pm$0.36} & 44.30{\scriptsize$\pm$0.43} \\
 & LC-CEM & \textbf{85.19}{\scriptsize$\pm$0.10} & 41.85{\scriptsize$\pm$1.22} & \textbf{2.03}{\scriptsize$\pm$0.34} & \textbf{53.34}{\scriptsize$\pm$0.83} \\
\bottomrule
\end{tabular}
}
\end{table*}

\paragraph{Concept embedding models.}
The latent-class structure is not specific to the CBM and
can also be added to other concept layers, such as Concept Embedding Models (CEM, \citep{espinosa2022concept}).
CEM represents each concept by an embedding vector rather than a single
probability.
For each concept $j$, it computes a ``present'' embedding
$\mathbf e_j^+(\bx)\in\R^m$ and an ``absent'' embedding
$\mathbf e_j^-(\bx)\in\R^m$, and mixes them according to the predicted
probability $p_j(\bx)$ that the concept is present,
\begin{equation}
\mathbf e_j(\bx)=p_j(\bx)\,\mathbf e_j^+(\bx)+\bigl(1-p_j(\bx)\bigr)\,\mathbf e_j^-(\bx),
\label{eq:cem}
\end{equation}
where $p_j(\bx)=\sigmoid\bigl(s_j(\bx)\bigr)$ and $s_j$ is a scoring function; the head then predicts $y$ from
$(\mathbf e_1,\dots,\mathbf e_C)$.
This improves target accuracy, but CEM still computes each $p_j$ separately and
thus treats the concepts as independent.
LC-CEM keeps the embeddings and the head of CEM and replaces only these
independent probabilities by a latent-class mixture,
\begin{equation}
p_j(\bx)=\sum_{k=1}^K\pi_k(\bx)\,\sigmoid\bigl(s_j(\bx)+b_{kj}\bigr),
\label{eq:lccem}
\end{equation}
so that the concepts are independent within each class but dependent overall.
Correcting one concept then updates $\pi_k$, and hence the probabilities and
embeddings of the other concepts.

\Cref{tab:cem} compares CEM and LC-CEM on CUB, CIFAR-10 and CIFAR-100 in concept accuracy, target accuracy, concept ECE and intervention AUC, with interventions following the uncertainty policy in Section \ref{sec:interven}.
LC-CEM improves concept accuracy, concept calibration and intervention AUC on all three datasets, with especially large gains in intervention AUC.
It also improves target accuracy on CUB and CIFAR-10, and matches CEM on CIFAR-100.
Implementation details are given in \cref{app:lccem}.

\begin{figure*}[t]
  \centering
  \includegraphics[width=\linewidth]{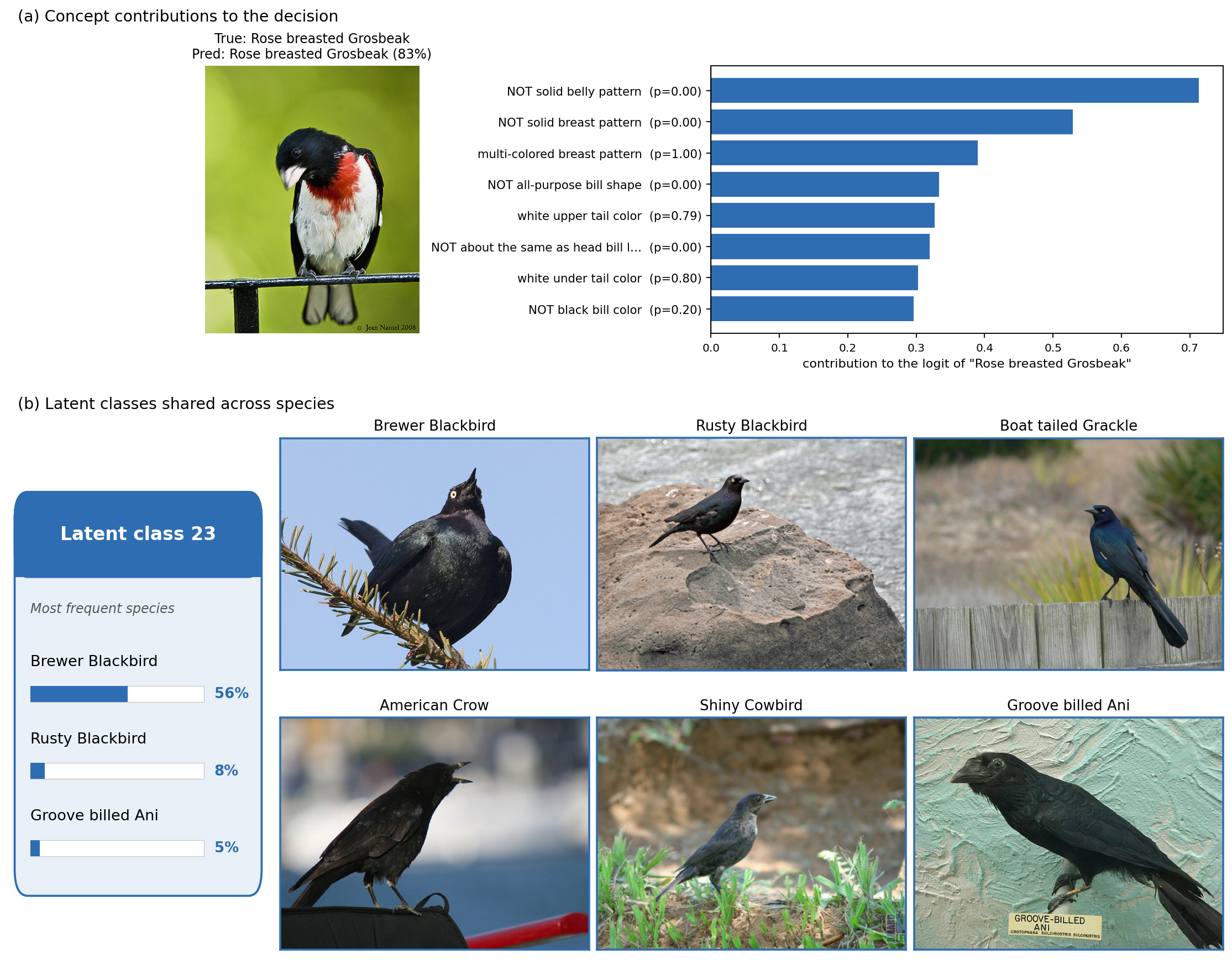}
  \caption{Interpreting LC-CBM on CUB.
  (a)~The eight largest concept contributions $W[y,j]\,(p(c_j{=}1\mid x) - \bar p_j)$ to the logit of the predicted species.
  Parentheses give $p(c_j{=}1\mid x)$.
  ``NOT'' marks concepts less likely than on average.
  (b)~Test images of different species whose most probable latent class is class~196.
  Left: the three most frequent species in this class over the train and test images.}
  \label{fig:interp}
\end{figure*}

\section{Interpretation}
\label{sec:interpretation}

We examine what a trained LC-CBM bases its decisions on and what its latent classes represent.
We use one LC-CBM trained on the CUB dataset.
\paragraph{Explaining a decision.}
Following \citet{oikarinen2023labelfree}, we explain a decision by how much each concept contributes to the logit of the predicted class.
The label predictor of LC-CBM is a linear layer with weights $W \in \mathbb{R}^{P \times C}$ and bias $W_0 \in \mathbb{R}^{P}$.
Given a concept vector $\mathbf{c} \in \{0,1\}^C$, it assigns each class $y$ the logit
\begin{equation}
  \mathrm{logit}_y(\mathbf{c}) = W_0[y] + \sum_{j=1}^{C} W[y,j]\,c_j,
\end{equation}
and $p(y \mid \mathbf{c}) = \mathrm{softmax}_y\big(\mathrm{logit}(\mathbf{c})\big)$.
Since the logit is linear in $\mathbf{c}$, its expectation under the predicted concept distribution decomposes exactly over concepts:
\begin{align*}
  \mathbb{E}_{p(\mathbf{c}\mid x)}\big[\mathrm{logit}_y(\mathbf{c})\big]
  &= W_0[y] + \sum_j W[y,j]\,\bar p_j \nonumber \\
  &\quad + \sum_j W[y,j]\,\big(p(c_j{=}1\mid x) - \bar p_j\big).
  \label{eq:contrib}
\end{align*}
Here $p(c_j{=}1\mid x) = \sum_k \pi_k(x)\,p_{kj}(x)$ is the marginal probability of concept $j$, and $\bar p_j$ is its average over the training images.
The first two terms give the logit of an average image.
The last sum attributes the deviation from that average to the individual concepts.
We mark a concept ``NOT'' when it is less likely than on average, $p(c_j{=}1\mid x) < \bar p_j$; a positive contribution of such a concept means that its \emph{absence} supports the prediction.

Figure~\ref{fig:interp}(a) explains the prediction for a Rose-breasted Grosbeak.
The strongest evidence is that its belly and breast are \emph{not} solid in pattern, while its breast is multi-colored.
White upper and under tail add further evidence.
These concepts describe the bird's red patch on a white breast and its black-and-white tail, the field marks that identify the species.
Several of the top contributions are absent concepts.
% This is common on CUB, where concepts are annotated per species \citep{koh2020concept}.
A species is therefore characterized as much by the features it lacks as by those it has.

\vspace{-7pt}
\paragraph{Latent classes are visual profiles shared across species.}
The latent classes of LC-CBM receive no supervision; the model learns them only through the concept likelihood.
Figure~\ref{fig:interp}(b) shows images of different species whose most probable latent class is the same.
Besides Brewer's and Rusty Blackbirds, the class contains a Boat-tailed Grackle, a Shiny Cowbird, an American Crow, and a Groove-billed Ani.
These species come from three families, but all are uniformly black.
In concept space they are nearly indistinguishable, and LC-CBM represents them with one shared concept profile.
The latent classes thus group species by appearance rather than taxonomy.
They mark exactly the species among which a decision is hard, and among which revealing a single distinguishing concept is most informative.
Further results are in Appendix \ref{app:interpretation}.

% =============================================================================================

\section{Conclusion}
\label{sec:conclusion}
\vspace{-4pt}
We proposed Latent Class CBMs (LC-CBMs), which model concepts as an input-dependent
mixture of product-Bernoulli distributions, so that correcting one concept
updates all others.
Unlike SCBM, LC-CBM has an exact likelihood and exact Bayesian interventions.
In our experiments, this made interventions more effective than with SCBM,
helped LC-CBM handle user mistakes, and carried over to CEM.
Future work includes selecting the number of latent classes automatically, for example with nonparametric Bayesian mixtures, and applying LC-CBM to domains such as medical imaging, where clinical findings are strongly dependent and concept annotations are often incomplete.


\bibliography{references}

\begin{thebibliography}{24}
\providecommand{\natexlab}[1]{#1}
\providecommand{\url}[1]{\texttt{#1}}
\expandafter\ifx\csname urlstyle\endcsname\relax
  \providecommand{\doi}[1]{doi: #1}\else
  \providecommand{\doi}{doi: \begingroup \urlstyle{rm}\Url}\fi

\bibitem[Arthur \& Vassilvitskii(2007)Arthur and Vassilvitskii]{arthur2007kmeans}
Arthur, D. and Vassilvitskii, S.
\newblock k-means++: The advantages of careful seeding.
\newblock In \emph{Proceedings of the Eighteenth Annual ACM-SIAM Symposium on Discrete Algorithms (SODA)}, pp.\  1027--1035, 2007.

\bibitem[Bengio et~al.(2013)Bengio, L{\'e}onard, and Courville]{bengio2013estimating}
Bengio, Y., L{\'e}onard, N., and Courville, A.
\newblock Estimating or propagating gradients through stochastic neurons for conditional computation.
\newblock \emph{arXiv preprint arXiv:1308.3432}, 2013.

\bibitem[Dempster et~al.(1977)Dempster, Laird, and Rubin]{dempster1977maximum}
Dempster, A.~P., Laird, N.~M., and Rubin, D.~B.
\newblock Maximum likelihood from incomplete data via the em algorithm.
\newblock \emph{Journal of the royal statistical society: series B (methodological)}, 39\penalty0 (1):\penalty0 1--22, 1977.

\bibitem[Dunson \& Xing(2009)Dunson and Xing]{dunson2009nonparametric}
Dunson, D.~B. and Xing, C.
\newblock Nonparametric bayes modeling of multivariate categorical data.
\newblock \emph{Journal of the American Statistical Association}, 104\penalty0 (487):\penalty0 1042--1051, 2009.

\bibitem[Espinosa~Zarlenga et~al.(2022)Espinosa~Zarlenga, Barbiero, Ciravegna, Marra, Giannini, Diligenti, Shams, Precioso, Melacci, Weller, et~al.]{espinosa2022concept}
Espinosa~Zarlenga, M., Barbiero, P., Ciravegna, G., Marra, G., Giannini, F., Diligenti, M., Shams, Z., Precioso, F., Melacci, S., Weller, A., et~al.
\newblock Concept embedding models: Beyond the accuracy-explainability trade-off.
\newblock \emph{Advances in neural information processing systems}, 35:\penalty0 21400--21413, 2022.

\bibitem[Goodman(1974)]{goodman1974exploratory}
Goodman, L.~A.
\newblock Exploratory latent structure analysis using both identifiable and unidentifiable models.
\newblock \emph{Biometrika}, 61\penalty0 (2):\penalty0 215--231, 1974.

\bibitem[Havasi et~al.(2022)Havasi, Parbhoo, and Doshi-Velez]{havasi2022addressing}
Havasi, M., Parbhoo, S., and Doshi-Velez, F.
\newblock Addressing leakage in concept bottleneck models.
\newblock \emph{Advances in Neural Information Processing Systems}, 35:\penalty0 23386--23397, 2022.

\bibitem[Jang et~al.(2016)Jang, Gu, and Poole]{jang2016categorical}
Jang, E., Gu, S., and Poole, B.
\newblock Categorical reparameterization with gumbel-softmax.
\newblock \emph{arXiv preprint arXiv:1611.01144}, 2016.

\bibitem[Koh et~al.(2020)Koh, Nguyen, Tang, Mussmann, Pierson, Kim, and Liang]{koh2020concept}
Koh, P.~W., Nguyen, T., Tang, Y.~S., Mussmann, S., Pierson, E., Kim, B., and Liang, P.
\newblock Concept bottleneck models.
\newblock In \emph{International conference on machine learning}, pp.\  5338--5348. Pmlr, 2020.

\bibitem[Krizhevsky(2009)]{krizhevsky2009learning}
Krizhevsky, A.
\newblock Learning multiple layers of features from tiny images.
\newblock Technical report, University of Toronto, 2009.

\bibitem[Mahinpei et~al.(2021)Mahinpei, Clark, Lage, Doshi-Velez, and Pan]{mahinpei2021promises}
Mahinpei, A., Clark, J., Lage, I., Doshi-Velez, F., and Pan, W.
\newblock Promises and pitfalls of black-box concept learning models.
\newblock \emph{arXiv preprint arXiv:2106.13314}, 2021.

\bibitem[Margeloiu et~al.(2021)Margeloiu, Ashman, Bhatt, Chen, Jamnik, and Weller]{margeloiu2021concept}
Margeloiu, A., Ashman, M., Bhatt, U., Chen, Y., Jamnik, M., and Weller, A.
\newblock Do concept bottleneck models learn as intended?
\newblock \emph{arXiv preprint arXiv:2105.04289}, 2021.

\bibitem[Naeini et~al.(2015)Naeini, Cooper, and Hauskrecht]{naeini2015obtaining}
Naeini, M.~P., Cooper, G., and Hauskrecht, M.
\newblock Obtaining well calibrated probabilities using bayesian binning.
\newblock In \emph{Proceedings of the AAAI conference on artificial intelligence}, volume~29, 2015.

\bibitem[Oikarinen et~al.(2023)Oikarinen, Das, Nguyen, and Weng]{oikarinen2023labelfree}
Oikarinen, T., Das, S., Nguyen, L.~M., and Weng, T.-W.
\newblock Label-free concept bottleneck models.
\newblock In \emph{International Conference on Learning Representations (ICLR)}, 2023.

\bibitem[Radford et~al.(2021)Radford, Kim, Hallacy, Ramesh, Goh, Agarwal, Sastry, Askell, Mishkin, Clark, Krueger, and Sutskever]{radford2021learning}
Radford, A., Kim, J.~W., Hallacy, C., Ramesh, A., Goh, G., Agarwal, S., Sastry, G., Askell, A., Mishkin, P., Clark, J., Krueger, G., and Sutskever, I.
\newblock Learning transferable visual models from natural language supervision.
\newblock In \emph{Proceedings of the 38th International Conference on Machine Learning (ICML)}, 2021.

\bibitem[Rao et~al.(2024)Rao, Mahajan, B{\"o}hle, and Schiele]{rao2024discover}
Rao, S., Mahajan, S., B{\"o}hle, M., and Schiele, B.
\newblock Discover-then-name: Task-agnostic concept bottlenecks via automated concept discovery.
\newblock In \emph{European Conference on Computer Vision}, pp.\  444--461. Springer, 2024.

\bibitem[Shang et~al.(2024)Shang, Zhou, Zhang, Ni, Yang, and Wang]{shang2024incremental}
Shang, C., Zhou, S., Zhang, H., Ni, X., Yang, Y., and Wang, Y.
\newblock Incremental residual concept bottleneck models.
\newblock In \emph{2024 IEEE/CVF Conference on Computer Vision and Pattern Recognition (CVPR)}, pp.\  11030--11040. IEEE, 2024.

\bibitem[Shin et~al.(2023)Shin, Jo, Ahn, and Lee]{shin2023closer}
Shin, S., Jo, Y., Ahn, S., and Lee, N.
\newblock A closer look at the intervention procedure of concept bottleneck models.
\newblock In \emph{International Conference on Machine Learning}, pp.\  31504--31520. PMLR, 2023.

\bibitem[Srivastava et~al.(2024)Srivastava, Yan, and Weng]{srivastava2024vlg}
Srivastava, D., Yan, G., and Weng, T.-W.
\newblock Vlg-cbm: Training concept bottleneck models with vision-language guidance.
\newblock \emph{Advances in Neural Information Processing Systems}, 37:\penalty0 79057--79094, 2024.

\bibitem[Vandenhirtz et~al.(2024)Vandenhirtz, Laguna, Marcinkevi{\v{c}}s, and Vogt]{vandenhirtz2024stochastic}
Vandenhirtz, M., Laguna, S., Marcinkevi{\v{c}}s, R., and Vogt, J.~E.
\newblock Stochastic concept bottleneck models.
\newblock \emph{Advances in Neural Information Processing Systems}, 37:\penalty0 51787--51810, 2024.

\bibitem[Wah et~al.(2011)Wah, Branson, Welinder, Perona, and Belongie]{wah2011cub}
Wah, C., Branson, S., Welinder, P., Perona, P., and Belongie, S.
\newblock The {Caltech-UCSD} birds-200-2011 dataset.
\newblock Technical Report CNS-TR-2011-001, California Institute of Technology, 2011.

\bibitem[Xu et~al.(2024)Xu, Qin, Mi, Wang, and Li]{xu2024energy}
Xu, X., Qin, Y., Mi, L., Wang, H., and Li, X.
\newblock Energy-based concept bottleneck models: Unifying prediction, concept intervention, and probabilistic interpretations.
\newblock In \emph{International Conference on Learning Representations}, volume 2024, pp.\  6944--6966, 2024.

\bibitem[Yuksekgonul et~al.(2022)Yuksekgonul, Wang, and Zou]{yuksekgonul2022post}
Yuksekgonul, M., Wang, M., and Zou, J.
\newblock Post-hoc concept bottleneck models.
\newblock \emph{arXiv preprint arXiv:2205.15480}, 2022.

\bibitem[Yun et~al.(2022)Yun, Bhalla, Pavlick, and Sun]{yun2022vision}
Yun, T., Bhalla, U., Pavlick, E., and Sun, C.
\newblock Do vision-language pretrained models learn composable primitive concepts?
\newblock \emph{arXiv preprint arXiv:2203.17271}, 2022.

\end{thebibliography}
\bibliographystyle{icml2027}

%%%%%%%%%%%%%%%%%%%%%%%%%%%%%%%%%%%%%%%%%%%%%%%%%%%%%%%%%%%%%%%%%%%%%%%%%%%%%%%
%%%%%%%%%%%%%%%%%%%%%%%%%%%%%%%%%%%%%%%%%%%%%%%%%%%%%%%%%%%%%%%%%%%%%%%%%%%%%%%
% APPENDIX
%%%%%%%%%%%%%%%%%%%%%%%%%%%%%%%%%%%%%%%%%%%%%%%%%%%%%%%%%%%%%%%%%%%%%%%%%%%%%%%
%%%%%%%%%%%%%%%%%%%%%%%%%%%%%%%%%%%%%%%%%%%%%%%%%%%%%%%%%%%%%%%%%%%%%%%%%%%%%%%
\newpage
\appendix
\onecolumn

% =============================================================================================
\section{Proofs}

\subsection{Derivation of the prediction in (\ref{eq:predict})}
\label{app:predict}
The target depends on the image only through the concepts, so
\begin{align}
p(y\mid\bx)
&=\sum_{\bc\in\{0,1\}^C}p_\phi(y\mid\bc)\,p_\theta(\bc\mid\bx) \nonumber\\
&=\sum_{\bc\in\{0,1\}^C}p_\phi(y\mid\bc)\sum_{k=1}^K\pi_k(\bx)\prod_{j=1}^C\Bern\bigl(c_j;p_{kj}(\bx)\bigr) \nonumber\\
&=\sum_{k=1}^K\pi_k(\bx)\underbrace{\sum_{\bc\in\{0,1\}^C}p_\phi(y\mid\bc)\prod_{j=1}^C\Bern\bigl(c_j;p_{kj}(\bx)\bigr)}_{=\;\E_{\bc\sim\prod_j\Bern(p_{kj}(\bx))}[p_\phi(y\mid\bc)]}
\end{align}

\qed

\subsection{Proof of \cref{prop:closure}}
\label{app:proofs}
\begin{proof}
Write $\Bern(c;p)=p^{c}(1-p)^{1-c}$ for $c\in\{0,1\}$. By \eqref{eq:lc}, the
model can be written with the latent class $z$ as
\begin{equation}
p(z=k,\bc\mid\bx)=\pi_k(\bx)\prod_{j=1}^C\Bern\bigl(c_j;p_{kj}(\bx)\bigr),
\label{eq:proof-joint}
\end{equation}
so that summing over $k$ recovers $p(\bc\mid\bx)$. Since $\cS$ and $\Sbar$
partition $\{1,\dots,C\}$, the product factorizes as
\begin{equation}
\prod_{j=1}^C\Bern\bigl(c_j;p_{kj}(\bx)\bigr)
=L_k(\bx,\bc_\cS)\prod_{j\in\Sbar}\Bern\bigl(c_j;p_{kj}(\bx)\bigr).
\label{eq:proof-factor}
\end{equation}

\emph{Step 1: likelihood of the revealed concepts.}
Summing \eqref{eq:proof-joint} over $\bc_\Sbar\in\{0,1\}^{|\Sbar|}$ and using
\eqref{eq:proof-factor},
\begin{equation*}
p(z=k,\bc_\cS\mid\bx)
=\pi_k(\bx)\,L_k(\bx,\bc_\cS)\prod_{j\in\Sbar}\;\sum_{c_j\in\{0,1\}}\Bern\bigl(c_j;p_{kj}(\bx)\bigr)
=\pi_k(\bx)\,L_k(\bx,\bc_\cS),
\end{equation*}
because the sum over $\bc_\Sbar$ of a product of functions of the individual
$c_j$ equals the product of the individual sums, each of which is $1$. Hence
\begin{equation*}
p(\bc_\cS\mid\bx)=\sum_{l=1}^K\pi_l(\bx)\,L_l(\bx,\bc_\cS).
\end{equation*}
This quantity is strictly positive, since $\pi_l(\bx)>0$ (softmax) and
$p_{lj}(\bx)\in(0,1)$ (sigmoid), so all conditional distributions below are
well defined.

\emph{Step 2: posterior class weights.}
By Bayes' rule and Step 1,
\begin{equation*}
p(z=k\mid\bx,\bc_\cS)
=\frac{p(z=k,\bc_\cS\mid\bx)}{p(\bc_\cS\mid\bx)}
=\frac{\pi_k(\bx)\,L_k(\bx,\bc_\cS)}{\sum_{l=1}^K\pi_l(\bx)\,L_l(\bx,\bc_\cS)}
=\pi_k(\bx,\bc_\cS),
\end{equation*}
which is \eqref{eq:posterior}.

\emph{Step 3: conditional distribution of the unrevealed concepts.}
By \eqref{eq:proof-joint} and \eqref{eq:proof-factor},
\begin{equation*}
p(\bc_\Sbar\mid\bx,\bc_\cS)
=\frac{\sum_{k=1}^K p(z=k,\bc\mid\bx)}{p(\bc_\cS\mid\bx)}
=\sum_{k=1}^K\frac{\pi_k(\bx)\,L_k(\bx,\bc_\cS)}{p(\bc_\cS\mid\bx)}
\prod_{j\in\Sbar}\Bern\bigl(c_j;p_{kj}(\bx)\bigr)
=\sum_{k=1}^K\pi_k(\bx,\bc_\cS)\prod_{j\in\Sbar}\Bern\bigl(c_j;p_{kj}(\bx)\bigr).
\end{equation*}
This is a latent class model of the form \eqref{eq:lc} over $\bc_\Sbar$ with
class weights $\pi_k(\bx,\bc_\cS)$ and the unchanged class-conditional
probabilities $p_{kj}(\bx)$, $j\in\Sbar$.

Finally, for $j\in\Sbar$, summing this distribution over all
$c_{j'}$ with $j'\in\Sbar\setminus\{j\}$ removes their factors as in Step 1,
which leaves
$p(c_j\mid\bx,\bc_\cS)=\sum_{k}\pi_k(\bx,\bc_\cS)\,\Bern\bigl(c_j;p_{kj}(\bx)\bigr)$.
Setting $c_j=1$ gives \eqref{eq:post-marginal}.
\end{proof}

\newpage

% =============================================================================================

\section{Details of the experiments}
\label{app:details}

\subsection{Data description}
\label{app:data}
\Cref{tab:data} summarizes the datasets used in our experiments. For CIFAR-10 and CIFAR-100, we use the concept labels of \citet{vandenhirtz2024stochastic}, who annotate the GPT-generated concept sets of \citet{oikarinen2023labelfree} with CLIP.

\begin{table}[h]
\centering
\caption{Descriptions of the datasets.}
\label{tab:data}
\begin{tabular}{lrrrl}
\toprule
Dataset & Data size & \# concepts & \# classes & Concept labels \\
\midrule
Synthetic \citep{vandenhirtz2024stochastic} & 50{,}000 & 100 & 2 & Generated \\
CUB-200 \citep{wah2011cub} & 11{,}788 & 112 & 200 & Human annotations \\
CIFAR-10 \citep{krizhevsky2009learning} & 60{,}000 & 143 & 10 & CLIP \\
CIFAR-100 \citep{krizhevsky2009learning} & 60{,}000 & 892 & 100 & CLIP \\
\bottomrule
\end{tabular}
\end{table}

\subsection{Training details}
\label{app:training}
All models are optimized with Adam. 
Each training set is split into training and validation sets; the validation set is used for hyperparameter selection and early stopping, and the test set only for the evaluation. 
On CIFAR-10 and CIFAR-100, we use 45{,}000 training and 5{,}000 validation images and the official 10{,}000 test images.
On CUB, we follow \citet{koh2020concept} and hold out 20\% of the official training set as validation data, giving 4{,}796 training, 1{,}198 validation and 5{,}794 test images.
The synthetic dataset has 50{,}000 samples, regenerated for every seed and split 60/20/20 into training, validation and test sets as in \citet{vandenhirtz2024stochastic}.
All results are averaged over ten seeds, except for CIFAR-100 (five seeds) and the ablations (three or five seeds, as stated in each table).

\subsection{EM initialization}

With many latent classes, LC-CBM can train poorly when started from random values.
We therefore set good starting values before training.
First, we group the training images into $K$ latent classes using only their concept label vectors $\bc$, not the images themselves or the targets.
This is done with the EM algorithm \citep{dempster1977maximum}, started with $k$-means++
\citep{arthur2007kmeans}; missing concept labels are skipped.
Let $r_{ik}$ be the probability, estimated by EM, that training image $i$
belongs to class $k$. We set $a_k=\log\hat\pi_k$ and
$b_{kj}=\operatorname{logit}\hat p_{kj}$, where
$\hat\pi_k=\frac1n\sum_i r_{ik}$ is the share of images in class $k$ and
$\hat p_{kj}=\sum_i r_{ik}c_{ij}/\sum_i r_{ik}$ is the frequency of concept
$j$ within it.
The other weights, $W_\pi$, $W_\mu$ and $\bm\beta$, start near zero, so that at the beginning of training the model simply uses the classes found by EM;
it then learns from the images how to adjust them.

% In all experiments, LC-CBM uses the shift form \eqref{eq:shift}, EM
% initialization, and $M_k=10$ concept samples per latent class for the target
% term.

\subsection{Hyperparameter selection}
\label{app:hparams}

For every model and dataset, we select the hyperparameters by grid search and choose the configuration with the highest target accuracy on the validation set.

\paragraph{LC-CBM.}
For all datasets, we select the hyperparameters of LC-CBM on the validation
set by grid search over the following ranges:
\begin{itemize}
  \item number of latent classes: $K\in\{10,50,200\}$;
  \item weight of the concept loss $\lambda\in\{0.1,0.3,1\}$;
  \item weight of the auxiliary head loss : $\alpha\in\{0.1,0.3,1\}$;
  \item learning rate $\bb$: $\{10^{-4},10^{-3}\}$;
  \item dropout probability: $\pdrop\in
        \{0,0.2,0.5,0.7,0.9\}$.
\end{itemize}

\paragraph{Baseline models.}
For Soft CBM, Hard CBM, Global SCBM and Amortized SCBM, we use the same candidate values as for LC-CBM for the concept-loss weight $\lambda$, the dropout probability and the learning rate, and the candidate values given in the respective original papers for all other hyperparameters.
% For each model, we choose the configuration that performs best on the validation set.

% \paragraph{Frozen-backbone experiments.}
% All models use the frozen 2048-d features of an ImageNet-pretrained
% ResNet-50, followed by a trainable adapter to 256 dimensions, and are trained
% for 150 epochs. For all datasets, the search range is $K\in\{50,200\}$,
% $\pdrop\in\{0,0.5\}$ and $\lambda\in\{0,1\}$, with $\alpha=1$. The selected
% configurations are:
% \begin{itemize}
%   \item \textbf{CIFAR-10}: $K=200$, $\pdrop=0.5$, $\lambda=1$.
%   \item \textbf{CUB}: $K=200$, $\pdrop=0$, $\lambda=1$.
%   \item \textbf{CUB-CLIP}: $K=50$, $\pdrop=0$, $\lambda=1$.
%   \item \textbf{CIFAR-100}: $K=50$, $\pdrop=0$, $\lambda=1$.
% \end{itemize}

\subsection{Details of latent class concept embedding models}

\label{app:lccem}

\paragraph{Model.}
CEM follows the implementation in the SCBM code base, with 16-dimensional
concept embeddings and no training-time interventions. LC-CEM uses the same
architecture, but the concept probabilities come from the latent-class model,
$\operatorname{logit} p_{kj}(\bx)=s_j(\bx)+b_{kj}$, where $s_j$ is the CEM
concept score. The embeddings are mixed with the marginal probabilities
$\sum_k\pi_k(\bx)\,p_{kj}(\bx)$. At intervention time, the class posterior is updated exactly given the intervened concepts.

\paragraph{Hyperparameters.}
For CEM and LC-CEM, all hyperparameters are selected from the same candidate values as in Appendix \ref{app:hparams}.

\subsection{Computational resources}
\label{app:compute}
All experiments were conducted using RTX 3090 and RTX 4090 GPUs with 24 GB VRAM, as well as RTX PRO 6000 GPUs with 96 GB VRAM.

% \subsection{Details for each experiment}
% \label{app:exp-details}

% % \paragraph{Interventions.}
% % Concepts are intervened one at a time by the uncertainty-based policy in \ref{sec:interven}, using the test concept labels as the user's answers.
% % The intervention AUC is the area under the target-accuracy curve divided by the range of the horizontal axis. \todo{Maximum number of
% % interventions per dataset.}

% \paragraph{Frozen backbones.}
% All models use the 2048-d features of an ImageNet-pretrained ResNet-50, which
% is kept fixed, followed by a trainable 256-d adapter; only the concept layer differs between models.

% \paragraph{CEM and LC-CEM.}
% We implement CEM as in the SCBM code, with embedding size 16 and without
% training-time interventions. The dropout rate and the auxiliary loss of
% LC-CEM are selected on validation data as for LC-CBM.

% \paragraph{Noisy interventions.}
% Each intervened value is flipped with probability $\varepsilon\in\{0.1,0.2\}$,
% with the same flips for all models; results use three seeds.

% \paragraph{Missing concept labels.}
% We remove 30\%, 50\% and 70\% of the training concept labels uniformly at
% random; the test labels are always complete.

% \paragraph{Label-free setting.}
% The setup, models and training of the label-free experiments are described
% in \cref{sec:lf_setup}.

% =============================================================================================

\newpage

\section{Robustness to noisy interventions and missing concept labels}
\label{sec:robustness}

\paragraph{Noisy interventions.}
In practice, users can make mistakes when they answer.
We simulate this by flipping each answer with probability $\varepsilon\in\{0.1,0.2\}$, using the same flips for every model.
The user therefore still answers with $0$ or $1$, but the answer is sometimes wrong.
We intervene with the uncertainty policy on CUB and CIFAR-10, and compare two ways in which a model can use the answers.
A \emph{naive} intervention trusts every answer and sets the concept to the given value.
A \emph{noise-aware} intervention takes into account that the answer is wrong with probability $\varepsilon$.
For Hard CBM and Amortized SCBM, it feeds $1-\varepsilon$ instead of $1$ (and $\varepsilon$ instead of $0$) to the model.
For LC-CBM, it replaces $p_{kj}$ in \eqref{eq:class-lik} with the probability that the user answers $1$ in class $k$,
\begin{equation}
p(\tilde c_j=1\mid z=k,\bx)=(1-\varepsilon)\,p_{kj}(\bx)+\varepsilon\,\bigl(1-p_{kj}(\bx)\bigr).
\label{eq:noisy}
\end{equation}
LC-CBM thus checks each answer against the other answers: a wrong answer that does not fit them changes the latent class weights only slightly.

\Cref{tab:noise} reports the intervention AUC over three seeds.
Noise-aware LC-CBM is the most robust model in all four settings; for example, on CUB with $\varepsilon=0.2$, its AUC is 93.7, whereas the other models reach at most 84.0.
For the baselines, the noise-aware intervention does not always help: it lowers the AUC of Hard CBM on CUB and of Amortized SCBM on CUB with $\varepsilon=0.1$.

\begin{table}[h]
\centering
\caption{Interventions with user errors: each revealed value is flipped with probability $\varepsilon$. AUC (unc.\ policy, three seeds).}
\label{tab:noise}
\begin{tabular}{llcccc}
\toprule
 & & \multicolumn{2}{c}{CUB} & \multicolumn{2}{c}{CIFAR-10} \\
\cmidrule(lr){3-4}\cmidrule(lr){5-6}
Model & Intervention & $\varepsilon=0.1$ & $\varepsilon=0.2$ & $\varepsilon=0.1$ & $\varepsilon=0.2$ \\
\midrule
\multirow{2}{*}{Hard CBM}
 & naive       & 92.93{\scriptsize$\pm$0.08} & 83.02{\scriptsize$\pm$0.07} & 75.06{\scriptsize$\pm$0.10} & 66.58{\scriptsize$\pm$0.21} \\
 & noise-aware & 83.37{\scriptsize$\pm$0.43} & 75.46{\scriptsize$\pm$0.13} & 78.50{\scriptsize$\pm$0.24} & 74.58{\scriptsize$\pm$0.28} \\
\midrule
\multirow{2}{*}{Amortized SCBM}
 & naive       & 94.20{\scriptsize$\pm$0.09} & 82.71{\scriptsize$\pm$0.29} & 75.91{\scriptsize$\pm$0.10} & 67.10{\scriptsize$\pm$0.09} \\
 & noise-aware & 91.14{\scriptsize$\pm$0.39} & 83.98{\scriptsize$\pm$0.61} & 80.28{\scriptsize$\pm$0.19} & 76.85{\scriptsize$\pm$0.21} \\
\midrule
\multirow{2}{*}{LC-CBM (ours)}
 & naive       & 95.85{\scriptsize$\pm$0.03} & 82.32{\scriptsize$\pm$0.17} & 76.84{\scriptsize$\pm$0.11} & 67.29{\scriptsize$\pm$0.07} \\
 & noise-aware & \textbf{97.41{\scriptsize$\pm$0.04}} & \textbf{93.71{\scriptsize$\pm$0.13}} & \textbf{81.52{\scriptsize$\pm$0.02}} & \textbf{78.50{\scriptsize$\pm$0.07}} \\
\bottomrule
\end{tabular}

\end{table}

\newpage

\paragraph{Missing concept labels.}
Concept labels are costly to collect, so in practice many of them are missing.
LC-CBM can be trained in this case by simply leaving the missing labels out of
its likelihood, and the baselines can likewise drop them from their concept
loss.
We compare the models in this setting with the same image backbone as in
\cref{sec:predict}, removing 30--70\% of the training concept labels at random.
As shown in \cref{tab:missing}, LC-CBM is the most accurate model in most settings; with 70\% of the labels missing, Amortized SCBM is better in target accuracy on CUB and both SCBMs are better in AUC on CIFAR-100.
Results under further settings are given in \cref{sec:lf}.

\begin{table}[h]
\centering
\caption{Target accuracy and intervention AUC (\%) under missing training concept labels (30/50/70\%, MCAR). Same backbones as the main experiments; mean $\pm$ std.\ over three seeds.}
\label{tab:missing}
\begin{tabular}{llccccc}
\toprule
Dataset & Missing & \begin{tabular}[c]{@{}c@{}}Soft\\CBM\end{tabular} & \begin{tabular}[c]{@{}c@{}}Hard\\CBM\end{tabular} & \begin{tabular}[c]{@{}c@{}}Global\\SCBM\end{tabular} & \begin{tabular}[c]{@{}c@{}}Amortized\\SCBM\end{tabular} & \begin{tabular}[c]{@{}c@{}}\textbf{LC-CBM}\\\textbf{(ours)}\end{tabular} \\
\midrule
\multicolumn{7}{l}{\textit{Target accuracy (\%) $\uparrow$}} \\
\midrule
\multirow{3}{*}{CUB} & 30\% & 67.4{\tiny$\pm$0.6} & 67.6{\tiny$\pm$0.4} & 67.4{\tiny$\pm$0.7} & 69.7{\tiny$\pm$0.4} & \textbf{70.9}{\tiny$\pm$0.9} \\
 & 50\% & 66.6{\tiny$\pm$0.1} & 66.7{\tiny$\pm$0.4} & 66.5{\tiny$\pm$0.5} & 69.1{\tiny$\pm$0.1} & \textbf{70.8}{\tiny$\pm$0.5} \\
 & 70\% & 65.3{\tiny$\pm$1.0} & 65.5{\tiny$\pm$0.7} & 65.4{\tiny$\pm$0.3} & \textbf{67.3}{\tiny$\pm$0.2} & 65.6{\tiny$\pm$0.5} \\
\cmidrule(l){1-7}
\multirow{3}{*}{CIFAR-10} & 30\% & 69.2{\tiny$\pm$0.2} & 68.3{\tiny$\pm$0.6} & 69.5{\tiny$\pm$0.6} & 70.1{\tiny$\pm$0.4} & \textbf{71.4}{\tiny$\pm$0.1} \\
 & 50\% & 68.3{\tiny$\pm$0.7} & 67.9{\tiny$\pm$0.2} & 68.4{\tiny$\pm$0.4} & 69.8{\tiny$\pm$0.2} & \textbf{71.7}{\tiny$\pm$0.5} \\
 & 70\% & 67.3{\tiny$\pm$0.5} & 66.5{\tiny$\pm$0.6} & 67.5{\tiny$\pm$0.2} & 68.3{\tiny$\pm$0.5} & \textbf{70.9}{\tiny$\pm$0.1} \\
\cmidrule(l){1-7}
\multirow{3}{*}{CIFAR-100} & 30\% & 39.1{\tiny$\pm$0.5} & 38.1{\tiny$\pm$0.9} & 39.9{\tiny$\pm$0.6} & 38.0{\tiny$\pm$0.4} & \textbf{41.0}{\tiny$\pm$0.5} \\
 & 50\% & 38.9{\tiny$\pm$0.5} & 38.4{\tiny$\pm$0.2} & 39.5{\tiny$\pm$0.4} & 37.5{\tiny$\pm$0.5} & \textbf{40.7}{\tiny$\pm$0.5} \\
 & 70\% & 38.6{\tiny$\pm$0.5} & 37.7{\tiny$\pm$0.5} & 39.2{\tiny$\pm$0.1} & 37.5{\tiny$\pm$1.0} & \textbf{40.7}{\tiny$\pm$0.4} \\
\midrule
\multicolumn{7}{l}{\textit{Intervention AUC (unc.) $\uparrow$}} \\
\midrule
\multirow{3}{*}{CUB} & 30\% & 95.9{\tiny$\pm$0.4} & 95.9{\tiny$\pm$0.4} & 97.2{\tiny$\pm$0.3} & 97.8{\tiny$\pm$0.0} & \textbf{99.2}{\tiny$\pm$0.0} \\
 & 50\% & 95.8{\tiny$\pm$0.3} & 95.9{\tiny$\pm$0.1} & 97.0{\tiny$\pm$0.4} & 97.5{\tiny$\pm$0.1} & \textbf{99.1}{\tiny$\pm$0.0} \\
 & 70\% & 94.7{\tiny$\pm$0.4} & 94.8{\tiny$\pm$0.1} & 95.7{\tiny$\pm$0.4} & 96.2{\tiny$\pm$0.8} & \textbf{97.1}{\tiny$\pm$0.4} \\
\cmidrule(l){1-7}
\multirow{3}{*}{CIFAR-10} & 30\% & 81.3{\tiny$\pm$0.1} & 81.7{\tiny$\pm$0.1} & 82.3{\tiny$\pm$0.0} & 82.7{\tiny$\pm$0.1} & \textbf{83.9}{\tiny$\pm$0.1} \\
 & 50\% & 81.2{\tiny$\pm$0.3} & 81.3{\tiny$\pm$0.1} & 81.8{\tiny$\pm$0.1} & 82.2{\tiny$\pm$0.2} & \textbf{83.3}{\tiny$\pm$0.1} \\
 & 70\% & 80.4{\tiny$\pm$0.1} & 80.7{\tiny$\pm$0.2} & 81.2{\tiny$\pm$0.1} & 81.4{\tiny$\pm$0.2} & \textbf{82.4}{\tiny$\pm$0.1} \\
\cmidrule(l){1-7}
\multirow{3}{*}{CIFAR-100} & 30\% & 45.8{\tiny$\pm$0.5} & 45.0{\tiny$\pm$0.6} & 47.6{\tiny$\pm$0.3} & 47.9{\tiny$\pm$0.3} & \textbf{49.9}{\tiny$\pm$0.0} \\
 & 50\% & 45.8{\tiny$\pm$0.4} & 45.3{\tiny$\pm$0.2} & 47.0{\tiny$\pm$0.3} & 47.3{\tiny$\pm$0.1} & \textbf{48.3}{\tiny$\pm$0.1} \\
 & 70\% & 44.9{\tiny$\pm$0.3} & 44.2{\tiny$\pm$0.5} & \textbf{46.4}{\tiny$\pm$0.5} & 45.9{\tiny$\pm$1.3} & 45.3{\tiny$\pm$0.4} \\
\bottomrule
\end{tabular}

\end{table}

\newpage

\section{Label-free, incomplete concept supervision}
\label{sec:lf}
% All numbers in tables/lf/*.tex are generated from results/lf/eval/*.json by src/lf/make_tex.py.

Label-free CBMs use concept labels generated by a vision--language model (here, CLIP) instead of human annotations. 
These labels are noisy, and in practice often only partly available.
We test whether LC-CBM, which models the joint distribution of concepts, remains effective in this setting, compared with Hard CBM and LF-CBM \citep{oikarinen2023labelfree}.
Unlike the main experiments, the backbone is a frozen CLIP encoder and far fewer labels are available.

\paragraph{Incomplete supervision.}
We randomly remove concept labels from the training and validation sets as follows:
\begin{itemize}
\item \emph{Complete}: no labels are removed.
\item \emph{Random labels} ($\rho\in\{0.3,0.1,0.03\}$): each label of each image is kept with probability $\rho$ and removed otherwise. For example, $\rho=0.1$ keeps 10\% of all labels.
\item \emph{Random images}: all concept labels are kept for a random 10\% of the images, and all concept labels of the other 90\% are removed.
\end{itemize}
Test labels are never removed.
In interventions, they are the true values that the user reveals.

\paragraph{Models.}
All concept models share an encoder $h(\bx)=\mathrm{ReLU}(W\,\mathrm{dropout}(\bx))$ with 512 units on top of the frozen embedding, and a linear task head on hard (0/1) concept vectors.
We compare LC-CBM ($K=200$ for CIFAR-10 and $K=100$ for CIFAR-100), Hard CBM, and Label-free CBM (LF-CBM).
For LF-CBM, we use the \href{https://github.com/Trustworthy-ML-Lab/Label-free-CBM}{official implementation} of \citet{oikarinen2023labelfree}.

\paragraph{Metrics.}
\begin{itemize}
\item \textbf{Acc@$n$}: test accuracy after intervening on $n$ concepts in a random order.
\item \textbf{Int-AUC}: area under the Acc@$n$ curve, with $n$ normalized to $[0,1]$, defined as in Section~\ref{sec:interven} but with concepts revealed in a random order.
\item \textbf{Imp-AUROC}: AUROC of the predicted probabilities of the concepts that remain hidden after intervening on 20 concepts.
\item \textbf{Joint NLL}: $-\log p(\bc\mid\bx)$ of the complete test concept vector, where lower is better.
\end{itemize}
All results are reported as mean $\pm$ standard deviation over three seeds.

\begin{figure}[h]
\centering
\includegraphics[width=1.0\textwidth]{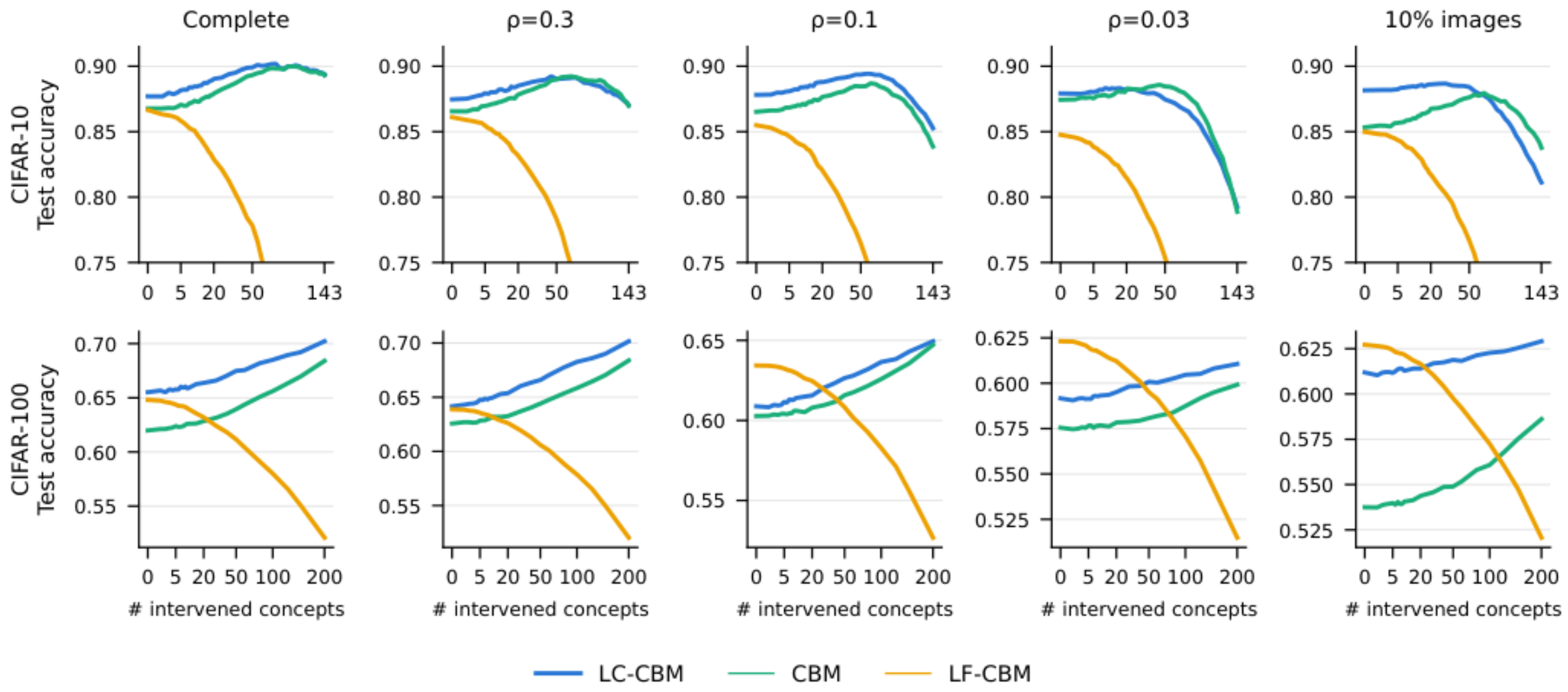}
\captionof{figure}{Test accuracy as the number of intervened concepts increases (mean over 3 seeds). The upper row presents CIFAR-10 and the lower row CIFAR-100.}
\label{fig:lf_int}
\end{figure}

\paragraph{Results.}
\begin{itemize}
    \item \textbf{CIFAR-10.} The upper part of \Cref{tab:lf_main} shows the results on CIFAR-10 with incomplete concept labels.
LC-CBM has the highest accuracy before intervention and after 10 interventions in every label setting, as well as the best joint NLL and Imp-AUROC, so it stays accurate even when most concept labels are missing.
The only exceptions are the later intervention steps with few labels: Hard CBM ties LC-CBM in Acc@50 and Int-AUC at $\rho=0.3$, is better in both at $\rho=0.03$, and has a slightly higher Int-AUC with 10\% images.
LF-CBM is below the others and its accuracy decreases as more concepts are revealed.
\item \textbf{CIFAR-100.} The lower part of \Cref{tab:lf_main} shows the results on CIFAR-100.
LC-CBM is best on every metric with complete labels and at $\rho=0.3$.
With fewer labels, LF-CBM has the highest accuracy before intervention (0.623--0.634 vs.\ 0.592--0.612 for LC-CBM), but its accuracy decreases as concepts are revealed while LC-CBM's increases. LC-CBM therefore has the best Acc@50, Int-AUC, Imp-AUROC and joint NLL in all five settings.
\item \textbf{Intervention curves.}
\Cref{fig:lf_int} presents the test accuracy as the number of intervened concepts increases.
On CIFAR-100, the accuracy of LC-CBM increases steadily with the number of interventions and is the highest or tied for the highest at $n=200$ in every setting. In contrast, the accuracy of LF-CBM decreases, so it falls below LC-CBM after a few interventions even where it starts higher (sparse labels).
On CIFAR-10, LC-CBM is the best or close to the best for most of the curve. Hard CBM catches up as more concepts are revealed, and with sparse labels it eventually overtakes LC-CBM (after about 50 interventions at $\rho=0.03$); the accuracy of both models drops when nearly all 143 concepts are intervened.
\end{itemize}

\newpage

\begin{table*}[!t]
\centering
\small
\setlength{\tabcolsep}{3pt}
\caption{Results with incomplete concept labels (mean$\pm$std over 3 seeds). Bold: best per column within each label setting.}
\label{tab:lf_main}
\resizebox{\textwidth}{!}{%
\begin{tabular}{llcccccc}
\toprule
Labels & Model & Acc@0 & Acc@10 & Acc@50 & Int-AUC & Imp-AUROC & Joint NLL$\downarrow$ \\
\midrule
\multicolumn{8}{c}{\textit{CIFAR-10}} \\
\midrule
\multirow{3}{*}{Complete} & LC-CBM (ours) & \textbf{0.877{\scriptsize$\pm$0.002}} & \textbf{0.885{\scriptsize$\pm$0.001}} & \textbf{0.899{\scriptsize$\pm$0.001}} & \textbf{0.896{\scriptsize$\pm$0.001}} & \textbf{0.961{\scriptsize$\pm$0.000}} & \textbf{30.2{\scriptsize$\pm$0.0}} \\
 & Hard CBM & 0.868{\scriptsize$\pm$0.003} & 0.874{\scriptsize$\pm$0.002} & 0.893{\scriptsize$\pm$0.001} & 0.892{\scriptsize$\pm$0.002} & 0.945{\scriptsize$\pm$0.000} & 40.8{\scriptsize$\pm$0.0} \\
 & LF-CBM & 0.867{\scriptsize$\pm$0.000} & 0.851{\scriptsize$\pm$0.001} & 0.778{\scriptsize$\pm$0.001} & 0.665{\scriptsize$\pm$0.003} & 0.884{\scriptsize$\pm$0.000} & -- \\
\midrule
\multirow{3}{*}{$\rho=0.3$} & LC-CBM (ours) & \textbf{0.875{\scriptsize$\pm$0.001}} & \textbf{0.882{\scriptsize$\pm$0.002}} & \textbf{0.890{\scriptsize$\pm$0.002}} & \textbf{0.884{\scriptsize$\pm$0.001}} & \textbf{0.956{\scriptsize$\pm$0.002}} & \textbf{32.6{\scriptsize$\pm$0.0}} \\
 & Hard CBM & 0.866{\scriptsize$\pm$0.004} & 0.873{\scriptsize$\pm$0.003} & \textbf{0.890{\scriptsize$\pm$0.003}} & \textbf{0.884{\scriptsize$\pm$0.004}} & 0.941{\scriptsize$\pm$0.000} & 42.1{\scriptsize$\pm$0.0} \\
 & LF-CBM & 0.861{\scriptsize$\pm$0.003} & 0.848{\scriptsize$\pm$0.002} & 0.783{\scriptsize$\pm$0.002} & 0.677{\scriptsize$\pm$0.004} & 0.881{\scriptsize$\pm$0.000} & -- \\
\midrule
\multirow{3}{*}{$\rho=0.1$} & LC-CBM (ours) & \textbf{0.878{\scriptsize$\pm$0.003}} & \textbf{0.884{\scriptsize$\pm$0.002}} & \textbf{0.894{\scriptsize$\pm$0.003}} & \textbf{0.883{\scriptsize$\pm$0.004}} & \textbf{0.944{\scriptsize$\pm$0.002}} & \textbf{37.1{\scriptsize$\pm$0.0}} \\
 & Hard CBM & 0.865{\scriptsize$\pm$0.001} & 0.871{\scriptsize$\pm$0.001} & 0.885{\scriptsize$\pm$0.002} & 0.873{\scriptsize$\pm$0.002} & 0.935{\scriptsize$\pm$0.000} & 44.9{\scriptsize$\pm$0.2} \\
 & LF-CBM & 0.855{\scriptsize$\pm$0.000} & 0.840{\scriptsize$\pm$0.001} & 0.765{\scriptsize$\pm$0.004} & 0.655{\scriptsize$\pm$0.000} & 0.878{\scriptsize$\pm$0.000} & -- \\
\midrule
\multirow{3}{*}{$\rho=0.03$} & LC-CBM (ours) & \textbf{0.879{\scriptsize$\pm$0.002}} & \textbf{0.883{\scriptsize$\pm$0.001}} & 0.875{\scriptsize$\pm$0.005} & 0.856{\scriptsize$\pm$0.004} & \textbf{0.933{\scriptsize$\pm$0.002}} & \textbf{45.5{\scriptsize$\pm$0.3}} \\
 & Hard CBM & 0.874{\scriptsize$\pm$0.002} & 0.877{\scriptsize$\pm$0.002} & \textbf{0.885{\scriptsize$\pm$0.002}} & \textbf{0.862{\scriptsize$\pm$0.002}} & 0.920{\scriptsize$\pm$0.000} & 51.9{\scriptsize$\pm$0.2} \\
 & LF-CBM & 0.848{\scriptsize$\pm$0.001} & 0.831{\scriptsize$\pm$0.003} & 0.754{\scriptsize$\pm$0.006} & 0.650{\scriptsize$\pm$0.003} & 0.876{\scriptsize$\pm$0.001} & -- \\
\midrule
\multirow{3}{*}{10\% images} & LC-CBM (ours) & \textbf{0.882{\scriptsize$\pm$0.001}} & \textbf{0.884{\scriptsize$\pm$0.001}} & \textbf{0.884{\scriptsize$\pm$0.008}} & 0.864{\scriptsize$\pm$0.003} & \textbf{0.939{\scriptsize$\pm$0.007}} & \textbf{36.2{\scriptsize$\pm$0.3}} \\
 & Hard CBM & 0.853{\scriptsize$\pm$0.009} & 0.859{\scriptsize$\pm$0.009} & 0.876{\scriptsize$\pm$0.009} & \textbf{0.866{\scriptsize$\pm$0.012}} & 0.932{\scriptsize$\pm$0.000} & 45.9{\scriptsize$\pm$0.4} \\
 & LF-CBM & 0.850{\scriptsize$\pm$0.001} & 0.837{\scriptsize$\pm$0.001} & 0.767{\scriptsize$\pm$0.006} & 0.662{\scriptsize$\pm$0.009} & 0.879{\scriptsize$\pm$0.000} & -- \\
\midrule
\multicolumn{8}{c}{\textit{CIFAR-100}} \\
\midrule
\multirow{3}{*}{Complete} & LC-CBM (ours) & \textbf{0.655{\scriptsize$\pm$0.001}} & \textbf{0.659{\scriptsize$\pm$0.002}} & \textbf{0.675{\scriptsize$\pm$0.002}} & \textbf{0.683{\scriptsize$\pm$0.002}} & \textbf{0.960{\scriptsize$\pm$0.002}} & \textbf{194.8{\scriptsize$\pm$0.1}} \\
 & Hard CBM & 0.620{\scriptsize$\pm$0.002} & 0.626{\scriptsize$\pm$0.002} & 0.640{\scriptsize$\pm$0.001} & 0.655{\scriptsize$\pm$0.001} & 0.944{\scriptsize$\pm$0.000} & 260.2{\scriptsize$\pm$0.5} \\
 & LF-CBM & 0.648{\scriptsize$\pm$0.001} & 0.640{\scriptsize$\pm$0.001} & 0.612{\scriptsize$\pm$0.002} & 0.581{\scriptsize$\pm$0.000} & 0.891{\scriptsize$\pm$0.000} & -- \\
\midrule
\multirow{3}{*}{$\rho=0.3$} & LC-CBM (ours) & \textbf{0.642{\scriptsize$\pm$0.000}} & \textbf{0.649{\scriptsize$\pm$0.001}} & \textbf{0.666{\scriptsize$\pm$0.002}} & \textbf{0.678{\scriptsize$\pm$0.001}} & \textbf{0.956{\scriptsize$\pm$0.001}} & \textbf{209.9{\scriptsize$\pm$0.1}} \\
 & Hard CBM & 0.626{\scriptsize$\pm$0.004} & 0.632{\scriptsize$\pm$0.003} & 0.644{\scriptsize$\pm$0.003} & 0.657{\scriptsize$\pm$0.004} & 0.941{\scriptsize$\pm$0.000} & 266.1{\scriptsize$\pm$0.2} \\
 & LF-CBM & 0.639{\scriptsize$\pm$0.003} & 0.632{\scriptsize$\pm$0.002} & 0.606{\scriptsize$\pm$0.002} & 0.579{\scriptsize$\pm$0.002} & 0.888{\scriptsize$\pm$0.000} & -- \\
\midrule
\multirow{3}{*}{$\rho=0.1$} & LC-CBM (ours) & 0.609{\scriptsize$\pm$0.007} & 0.614{\scriptsize$\pm$0.007} & \textbf{0.626{\scriptsize$\pm$0.009}} & \textbf{0.634{\scriptsize$\pm$0.009}} & \textbf{0.947{\scriptsize$\pm$0.003}} & \textbf{230.0{\scriptsize$\pm$0.1}} \\
 & Hard CBM & 0.603{\scriptsize$\pm$0.004} & 0.606{\scriptsize$\pm$0.005} & 0.616{\scriptsize$\pm$0.005} & 0.625{\scriptsize$\pm$0.005} & 0.937{\scriptsize$\pm$0.000} & 277.7{\scriptsize$\pm$0.3} \\
 & LF-CBM & \textbf{0.634{\scriptsize$\pm$0.002}} & \textbf{0.630{\scriptsize$\pm$0.001}} & 0.608{\scriptsize$\pm$0.000} & 0.582{\scriptsize$\pm$0.003} & 0.885{\scriptsize$\pm$0.000} & -- \\
\midrule
\multirow{3}{*}{$\rho=0.03$} & LC-CBM (ours) & 0.592{\scriptsize$\pm$0.004} & 0.593{\scriptsize$\pm$0.005} & \textbf{0.601{\scriptsize$\pm$0.005}} & \textbf{0.603{\scriptsize$\pm$0.005}} & \textbf{0.938{\scriptsize$\pm$0.002}} & \textbf{264.2{\scriptsize$\pm$0.2}} \\
 & Hard CBM & 0.576{\scriptsize$\pm$0.011} & 0.577{\scriptsize$\pm$0.011} & 0.581{\scriptsize$\pm$0.011} & 0.588{\scriptsize$\pm$0.010} & 0.927{\scriptsize$\pm$0.000} & 301.6{\scriptsize$\pm$1.2} \\
 & LF-CBM & \textbf{0.623{\scriptsize$\pm$0.002}} & \textbf{0.617{\scriptsize$\pm$0.002}} & 0.595{\scriptsize$\pm$0.002} & 0.569{\scriptsize$\pm$0.002} & 0.882{\scriptsize$\pm$0.000} & -- \\
\midrule
\multirow{3}{*}{10\% images} & LC-CBM (ours) & 0.612{\scriptsize$\pm$0.012} & 0.613{\scriptsize$\pm$0.012} & \textbf{0.619{\scriptsize$\pm$0.017}} & \textbf{0.622{\scriptsize$\pm$0.017}} & \textbf{0.933{\scriptsize$\pm$0.001}} & \textbf{236.3{\scriptsize$\pm$2.9}} \\
 & Hard CBM & 0.537{\scriptsize$\pm$0.012} & 0.541{\scriptsize$\pm$0.014} & 0.549{\scriptsize$\pm$0.013} & 0.562{\scriptsize$\pm$0.012} & 0.932{\scriptsize$\pm$0.000} & 288.4{\scriptsize$\pm$0.3} \\
 & LF-CBM & \textbf{0.627{\scriptsize$\pm$0.002}} & \textbf{0.622{\scriptsize$\pm$0.002}} & 0.598{\scriptsize$\pm$0.004} & 0.573{\scriptsize$\pm$0.003} & 0.885{\scriptsize$\pm$0.000} & -- \\
\bottomrule
\end{tabular}}
\vspace{1em}
\end{table*}

\newpage

$\quad$

\newpage

\section{Experiments with Different Intervention Policies}
\label{app:eig}

We also evaluate two further intervention policies. Under the \emph{random} policy, concepts are revealed in a uniformly random order. This removes any advantage that comes from choosing which concept to ask about, so it isolates how well each model propagates the revealed values. The \emph{expected information gain} (EIG) policy selects the concept whose true value is expected to reduce the entropy of the target prediction the most. LC-CBM's exact joint distribution makes EIG easy to compute, and we apply the same policy to all models. Table~\ref{tab:policies} shows that LC-CBM has the best intervention AUC under all three policies on all three datasets.

\begin{table}[h]
\centering
\footnotesize
\caption{Intervention policies: uncertainty vs.\ random vs.\ EIG (target-accuracy AUC and accuracy after 5 interventions, \%; mean $\pm$ std.\ over three seeds).}
\label{tab:policies}
\setlength{\tabcolsep}{3pt}
\begin{tabular}{llcccccc}
\toprule
 & & \multicolumn{3}{c}{AUC} & \multicolumn{3}{c}{Acc@5} \\
\cmidrule(lr){3-5}\cmidrule(lr){6-8}
Dataset & Model & Unc. & Random & EIG & Unc. & Random & EIG \\
\midrule
\multirow{3}{*}{Synthetic}
 & Hard CBM       & 69.12{\scriptsize$\pm$3.12} & 69.59{\scriptsize$\pm$3.06} & 74.09{\scriptsize$\pm$3.47} & 58.3{\scriptsize$\pm$2.6} & 58.0{\scriptsize$\pm$2.6} & 62.6{\scriptsize$\pm$1.9} \\
 & Amortized SCBM & 73.26{\scriptsize$\pm$3.73} & 72.41{\scriptsize$\pm$3.52} & 75.29{\scriptsize$\pm$3.75} & 62.0{\scriptsize$\pm$2.2} & 61.4{\scriptsize$\pm$2.0} & 68.6{\scriptsize$\pm$2.1} \\
 & LC-CBM         & \textbf{75.50}{\scriptsize$\pm$3.34} & \textbf{74.27}{\scriptsize$\pm$3.18} & \textbf{78.27}{\scriptsize$\pm$3.69} & \textbf{62.9}{\scriptsize$\pm$2.5} & \textbf{62.8}{\scriptsize$\pm$2.0} & \textbf{71.6}{\scriptsize$\pm$3.8} \\
\midrule
\multirow{3}{*}{CUB}
 & Hard CBM       & 96.03{\scriptsize$\pm$0.12} & 89.65{\scriptsize$\pm$0.32} & 96.24{\scriptsize$\pm$0.08} & 77.1{\scriptsize$\pm$0.8} & 69.5{\scriptsize$\pm$0.8} & 77.2{\scriptsize$\pm$0.7} \\
 & Amortized SCBM & 98.04{\scriptsize$\pm$0.06} & 93.62{\scriptsize$\pm$0.13} & 97.72{\scriptsize$\pm$0.08} & 87.9{\scriptsize$\pm$0.3} & 76.1{\scriptsize$\pm$0.2} & 84.4{\scriptsize$\pm$0.6} \\
 & LC-CBM         & \textbf{99.32}{\scriptsize$\pm$0.01} & \textbf{96.56}{\scriptsize$\pm$0.10} & \textbf{99.27}{\scriptsize$\pm$0.03} & \textbf{97.0}{\scriptsize$\pm$0.2} & \textbf{81.5}{\scriptsize$\pm$0.1} & \textbf{97.0}{\scriptsize$\pm$0.2} \\
\midrule
\multirow{3}{*}{CIFAR-10}
 & Hard CBM       & 81.85{\scriptsize$\pm$0.09} & 79.55{\scriptsize$\pm$0.17} & 83.27{\scriptsize$\pm$0.06} & 71.0{\scriptsize$\pm$0.1} & 70.0{\scriptsize$\pm$0.3} & 76.0{\scriptsize$\pm$0.5} \\
 & Amortized SCBM & 82.98{\scriptsize$\pm$0.03} & 80.22{\scriptsize$\pm$0.15} & 84.23{\scriptsize$\pm$0.06} & 71.4{\scriptsize$\pm$0.2} & 67.4{\scriptsize$\pm$0.1} & 78.7{\scriptsize$\pm$0.1} \\
 & LC-CBM         & \textbf{84.48}{\scriptsize$\pm$0.04} & \textbf{82.09}{\scriptsize$\pm$0.06} & \textbf{85.42}{\scriptsize$\pm$0.03} & \textbf{74.9}{\scriptsize$\pm$0.4} & \textbf{73.4}{\scriptsize$\pm$0.2} & \textbf{81.0}{\scriptsize$\pm$0.1} \\
\bottomrule
\end{tabular}
\end{table}

\newpage

\section{Ablation studies}

\subsection{Ablation on the number of latent classes $K$}
\label{sec:ablation_k}

We vary the number of latent classes $K\in\{1,10,50,100,200\}$ on CIFAR-10 and CIFAR-100, keeping all other hyperparameters fixed.
With $K=1$, the concepts are independent given $\bx$, so this setting removes the latent class structure.

Removing the latent classes clearly hurts: going from $K=10$ to $K=1$ increases the joint NLL (CIFAR-10: 36.3 $\to$ 44.0; CIFAR-100: 245.9 $\to$ 304.7) and decreases the intervention AUC (CIFAR-10: 84.0 $\to$ 83.0; CIFAR-100: 48.9 $\to$ 44.8).
Beyond $K=10$, larger $K$ still helps, but the gains diminish quickly.
On CIFAR-10, the results change little for $K\ge10$; on CIFAR-100, which has more concepts, most of the improvement is obtained by $K=50$, and increasing $K$ further improves the AUC only slightly (50.7 $\to$ 51.2).
Since the target loss uses $M$ concept samples for each of the $K$ latent classes, its cost grows linearly in $K$, so a moderate $K$ offers a good trade-off between performance and computational cost.

\begin{table}[h]
\centering
\caption{Sensitivity of LC-CBM to the number of latent classes $K$ (selected configuration otherwise
unchanged; mean $\pm$ std over three seeds). $K=1$ removes the latent class structure.
Joint NLL is $-\log p(\mathbf{c}\mid\mathbf{x})$ in nats. AUC (unc.) is computed over
$n=0,\dots,C$ revealed concepts on CIFAR-10 and $n=0,\dots,100$ on CIFAR-100.}
\label{tab:ablation_k}
\small
\begin{tabular}{llccccc}
\toprule
Data & $K$ & Concept acc.\ $\uparrow$ & Target acc.\ $\uparrow$ & Joint NLL $\downarrow$ & AUC (unc.) $\uparrow$ & Target ECE $\downarrow$ \\
\midrule
\multirow{5}{*}{CIFAR-10}
 & 1   & 86.19{\scriptsize$\pm$0.03} & 69.20{\scriptsize$\pm$0.12} & 44.04{\scriptsize$\pm$0.07} & 83.01{\scriptsize$\pm$0.09} & 10.22{\scriptsize$\pm$0.27} \\
 & 10  & 86.31{\scriptsize$\pm$0.02} & 70.81{\scriptsize$\pm$0.24} & 36.29{\scriptsize$\pm$0.03} & 84.01{\scriptsize$\pm$0.08} & 11.40{\scriptsize$\pm$0.57} \\
 & 50  & 86.40{\scriptsize$\pm$0.01} & 70.26{\scriptsize$\pm$0.42} & 34.27{\scriptsize$\pm$0.02} & 83.99{\scriptsize$\pm$0.03} & 9.18{\scriptsize$\pm$0.73} \\
 & 100 & \textbf{86.48}{\scriptsize$\pm$0.02} & \textbf{71.37}{\scriptsize$\pm$0.10} & 33.53{\scriptsize$\pm$0.05} & 84.29{\scriptsize$\pm$0.09} & 8.12{\scriptsize$\pm$0.23} \\
 & 200 & \textbf{86.48}{\scriptsize$\pm$0.02} & 71.24{\scriptsize$\pm$0.20} & \textbf{33.20}{\scriptsize$\pm$0.03} & \textbf{84.48}{\scriptsize$\pm$0.03} & \textbf{6.32}{\scriptsize$\pm$0.07} \\
\midrule
\multirow{5}{*}{CIFAR-100}
 & 1   & 84.69{\scriptsize$\pm$0.03} & 35.82{\scriptsize$\pm$0.29} & 304.68{\scriptsize$\pm$0.29} & 44.84{\scriptsize$\pm$0.41} & 4.73{\scriptsize$\pm$0.37} \\
 & 10  & 84.94{\scriptsize$\pm$0.03} & 38.57{\scriptsize$\pm$0.51} & 245.92{\scriptsize$\pm$0.46} & 48.94{\scriptsize$\pm$0.22} & 4.35{\scriptsize$\pm$0.50} \\
 & 50  & 85.01{\scriptsize$\pm$0.03} & 38.90{\scriptsize$\pm$0.23} & 233.07{\scriptsize$\pm$0.35} & 50.74{\scriptsize$\pm$0.17} & 5.42{\scriptsize$\pm$0.58} \\
 & 100 & \textbf{85.04}{\scriptsize$\pm$0.01} & 39.47{\scriptsize$\pm$0.25} & 228.94{\scriptsize$\pm$0.29} & 50.91{\scriptsize$\pm$0.38} & 5.42{\scriptsize$\pm$0.59} \\
 & 200 & 85.03{\scriptsize$\pm$0.03} & \textbf{40.15}{\scriptsize$\pm$0.27} & \textbf{224.72}{\scriptsize$\pm$0.26} & \textbf{51.17}{\scriptsize$\pm$0.08} & \textbf{3.48}{\scriptsize$\pm$0.25} \\
\bottomrule
\end{tabular}
\end{table}

\subsection{Computational cost experiments}
\label{app:cost}

Table~\ref{tab:cost_k10} compares LC-CBM with $K{=}10$ to Hard CBM and Amortized SCBM.
The parameters and training memory of LC-CBM stay within a few percent of those of Hard CBM for every number of concepts.
With $K=10$, its training step takes about 11\,ms for every $C$, about two to three times that of Hard CBM.
In contrast, the parameters of Amortized SCBM grow quadratically with $C$, reaching 104M at $C{=}900$, and its training step grows from 10\,ms to 68\,ms as $C$ increases from 100 to 900.
An intervention with LC-CBM takes 1--3\,ms, whereas Amortized SCBM takes 0.7--1.2\,s because it solves a constrained optimization for every sample.
Thus, LC-CBM models dependencies among concepts at nearly the memory and parameter cost of Hard CBM, which treats concepts independently.
It also remains orders of magnitude faster than SCBM at intervention time.

\begin{table}[h]
  \centering
  \caption{Computational cost of the concept layer for $C$ concepts and $P = 200$ classes, on 256-dimensional features.
  A training step is the forward pass, backward pass, and Adam update on a batch of 64 (median of 20 runs).
  Training memory is the peak additional GPU memory of one training step, excluding parameters and optimizer state.
  An intervention step reveals 10 concepts for 512 images and predicts the label from 100 samples.
  Measured on one NVIDIA RTX 3090.}
  \label{tab:cost_k10}
  \small
  \begin{tabular}{rlrrrr}
    \toprule
    $C$ & Model & Parameters & Train step (ms) & Train memory (MB) & Intervention (ms) \\
    \midrule
    \multirow{3}{*}{100}
      & Hard CBM              & 25.7K  & 5.1  & 37  & 0.8 \\
      & Amortized SCBM             & 1.32M  & 10.1 & 44  & 673.3 \\
      & LC-CBM ($K{=}10$) & 29.3K  & 11.0 & 38  & 1.1 \\
    \midrule
    \multirow{3}{*}{300}
      & Hard CBM              & 77.1K  & 3.4  & 74  & 1.0 \\
      & Amortized SCBM             & 11.7M  & 14.3 & 158 & 725.2 \\
      & LC-CBM ($K{=}10$) & 82.7K  & 11.0 & 75  & 1.4 \\
    \midrule
    \multirow{3}{*}{900}
      & Hard CBM              & 231K   & 6.0  & 221 & 1.7 \\
      & Amortized SCBM             & 104M   & 67.6 & 979 & 1189.4 \\
      & LC-CBM ($K{=}10$) & 243K   & 11.2 & 225 & 2.8 \\
    \bottomrule
  \end{tabular}
\end{table}

\subsection{Sensitivity to the loss weights}
\label{app:loss_weights}

We study how the two weights in the training objective (Eq.~\ref{eq:loss}) affect LC-CBM.
The weight $\lambda$ scales the concept loss, and the weight $\alpha$ scales the loss of the auxiliary head.
We train LC-CBM on CIFAR-10 with $K{=}200$ and dropout $0.5$.
We vary one weight at a time, with $\lambda \in \{0.1, 0.3, 1, 3\}$ and $\alpha \in \{0, 0.3, 1, 3\}$, and keep all other settings fixed.
Table~\ref{tab:loss_weights} reports the mean and standard deviation over three seeds.

The weight $\alpha$ has little effect on most metrics.
A larger $\alpha$ worsens the target ECE, and removing the auxiliary head ($\alpha{=}0$) lowers the intervention AUC.
In contrast, $\lambda$ affects the results more broadly.
A smaller $\lambda$ improves the target accuracy, target NLL, and intervention AUC, while the concept metrics degrade slightly at the smallest value.
A value of $\lambda$ between 0.1 and 0.3 gives a good balance between concept and target performance.

\begin{table}[h]
  \centering
  \caption{Sensitivity of LC-CBM to the concept loss weight $\lambda$ and the auxiliary head loss weight $\alpha$ on CIFAR-10 (mean $\pm$ standard deviation over three seeds).
  Joint NLL is $-\log p(\mathbf{c}\mid x)$.
  AUC is the mean target accuracy over the number of concepts revealed by the uncertainty policy.}
  \label{tab:loss_weights}
  \resizebox{\linewidth}{!}{%
  \begin{tabular}{cc ccccccc}
    \toprule
    $\lambda$ & $\alpha$ & Concept acc.\ $\uparrow$ & Target acc.\ $\uparrow$ & Joint NLL $\downarrow$ & Target NLL $\downarrow$ & Target ECE $\downarrow$ & Concept ECE $\downarrow$ & AUC $\uparrow$ \\
    \midrule
    0.1 & 1 & 86.11{\scriptsize$\pm$0.01} & 74.97{\scriptsize$\pm$0.20} & 33.96{\scriptsize$\pm$0.05} & 0.768{\scriptsize$\pm$0.005} & 5.64{\scriptsize$\pm$0.36} & 1.73{\scriptsize$\pm$0.03} & 85.13{\scriptsize$\pm$0.08} \\
    0.3 & 1 & 86.45{\scriptsize$\pm$0.01} & 74.19{\scriptsize$\pm$0.23} & 33.31{\scriptsize$\pm$0.02} & 0.769{\scriptsize$\pm$0.004} & 1.33{\scriptsize$\pm$0.25} & 1.08{\scriptsize$\pm$0.04} & 84.87{\scriptsize$\pm$0.08} \\
    1   & 1 & 86.48{\scriptsize$\pm$0.02} & 71.24{\scriptsize$\pm$0.20} & 33.20{\scriptsize$\pm$0.03} & 0.853{\scriptsize$\pm$0.005} & 6.32{\scriptsize$\pm$0.07} & 0.97{\scriptsize$\pm$0.06} & 84.48{\scriptsize$\pm$0.03} \\
    3   & 1 & 86.40{\scriptsize$\pm$0.03} & 69.26{\scriptsize$\pm$0.21} & 33.28{\scriptsize$\pm$0.02} & 0.922{\scriptsize$\pm$0.009} & 9.43{\scriptsize$\pm$0.53} & 1.03{\scriptsize$\pm$0.01} & 84.16{\scriptsize$\pm$0.03} \\
    \midrule
    1 & 0   & 86.49{\scriptsize$\pm$0.01} & 71.52{\scriptsize$\pm$0.06} & 33.20{\scriptsize$\pm$0.02} & 0.845{\scriptsize$\pm$0.007} & 5.26{\scriptsize$\pm$0.10} & 0.97{\scriptsize$\pm$0.07} & 82.80{\scriptsize$\pm$0.01} \\
    1 & 0.3 & 86.47{\scriptsize$\pm$0.02} & 71.38{\scriptsize$\pm$0.36} & 33.17{\scriptsize$\pm$0.03} & 0.847{\scriptsize$\pm$0.001} & 5.97{\scriptsize$\pm$0.61} & 0.98{\scriptsize$\pm$0.04} & 84.16{\scriptsize$\pm$0.03} \\
    1 & 1   & 86.48{\scriptsize$\pm$0.02} & 71.24{\scriptsize$\pm$0.20} & 33.20{\scriptsize$\pm$0.03} & 0.853{\scriptsize$\pm$0.005} & 6.32{\scriptsize$\pm$0.07} & 0.97{\scriptsize$\pm$0.06} & 84.48{\scriptsize$\pm$0.03} \\
    1 & 3   & 86.47{\scriptsize$\pm$0.02} & 71.61{\scriptsize$\pm$0.14} & 33.22{\scriptsize$\pm$0.02} & 0.856{\scriptsize$\pm$0.004} & 7.24{\scriptsize$\pm$0.22} & 0.99{\scriptsize$\pm$0.05} & 84.60{\scriptsize$\pm$0.05} \\
    \bottomrule
  \end{tabular}}
\end{table}

\subsection{Effect of the Auxiliary Head Loss}
\label{app:auxhead}

To check whether the gains of LC-CBM come from the auxiliary head loss, we
add it ($\alpha=1$) to Hard CBM and Amortized SCBM, and remove it
($\alpha=0$) from LC-CBM (\cref{tab:auxhead}). For every model, the loss
raises the intervention AUC under the uncertainty-based policy by one to two
points and leaves the other metrics unchanged. With the loss added to all
models, LC-CBM remains best on all metrics.

\begin{table}[h]
\centering
\small
\caption{Effect of the auxiliary head loss on CIFAR-10 (\%). Brier and ECE
are computed per concept. AUC is the area under the intervention curve with
the uncertainty-based policy. Mean $\pm$ std.\ over five seeds (three for
LC-CBM with $\alpha=0$).}
\label{tab:auxhead}
\begin{tabular}{lcccccc}
\toprule
Model & $\alpha$ & Concept acc.\,$\uparrow$ & Target acc.\,$\uparrow$ & Brier\,$\downarrow$ & ECE\,$\downarrow$ & AUC\,$\uparrow$ \\
\midrule
Hard CBM        & 0 & 85.41{\scriptsize$\pm$0.07} & 68.90{\scriptsize$\pm$0.41} & 10.52{\scriptsize$\pm$0.06} & 5.16{\scriptsize$\pm$0.18} & 81.89{\scriptsize$\pm$0.09} \\
                & 1 & 85.40{\scriptsize$\pm$0.04} & 69.03{\scriptsize$\pm$0.26} & 10.51{\scriptsize$\pm$0.04} & 5.12{\scriptsize$\pm$0.15} & 82.89{\scriptsize$\pm$0.10} \\
\midrule
Amortized SCBM  & 0 & 85.92{\scriptsize$\pm$0.02} & 70.95{\scriptsize$\pm$0.35} & 9.91{\scriptsize$\pm$0.02} & 2.32{\scriptsize$\pm$0.10} & 82.95{\scriptsize$\pm$0.05} \\
                & 1 & 85.92{\scriptsize$\pm$0.02} & 70.88{\scriptsize$\pm$0.31} & 9.91{\scriptsize$\pm$0.02} & 2.33{\scriptsize$\pm$0.12} & 83.97{\scriptsize$\pm$0.06} \\
\midrule
LC-CBM          & 0 & 86.48{\scriptsize$\pm$0.03} & 71.42{\scriptsize$\pm$0.11} & 9.48{\scriptsize$\pm$0.01} & 0.99{\scriptsize$\pm$0.07} & 82.68{\scriptsize$\pm$0.07} \\
                & 1 & \textbf{86.48}{\scriptsize$\pm$0.02} & \textbf{71.46}{\scriptsize$\pm$0.15} & \textbf{9.48}{\scriptsize$\pm$0.01} & \textbf{0.94}{\scriptsize$\pm$0.04} & \textbf{84.47}{\scriptsize$\pm$0.04} \\
\bottomrule
\end{tabular}
\end{table}

\newpage

\section{Comparison with Energy-based CBM}
\label{app:ecbm}
We compare LC-CBM with the Energy-based Concept Bottleneck Model (ECBM)~\citep{xu2024energy}, which also models dependencies between concepts. ECBM predicts the class and the concepts by minimizing learned energy functions at test time. We use the official implementation with the same image encoders and training protocol as in our main experiments. Results are averaged over ten seeds (five on CIFAR-100).
Table~\ref{tab:ecbm} shows the results. LC-CBM outperforms ECBM in concept accuracy, concept calibration and intervention on all three datasets.

\begin{table}[h]
\centering
\caption{Comparison with ECBM (mean $\pm$ std over five or ten seeds). AUC is the intervention AUC under the uncertainty policy.}
\label{tab:ecbm}
\begin{tabular}{llcccc}
\toprule
Dataset & Model & Concept acc.\ $\uparrow$ & Target acc.\ $\uparrow$ & Concept ECE $\downarrow$ & AUC (unc.) $\uparrow$ \\
\midrule
\multirow{2}{*}{CUB}
 & ECBM          & 95.28$\pm$0.02 & 69.66$\pm$0.33 & 3.12$\pm$0.02 & 86.27$\pm$0.14 \\
 & LC-CBM (ours) & \textbf{96.07$\pm$0.05} & \textbf{72.35$\pm$0.40} & \textbf{1.09$\pm$0.04} & \textbf{99.31$\pm$0.01} \\
\midrule
\multirow{2}{*}{CIFAR-10}
 & ECBM          & 85.84$\pm$0.05 & \textbf{71.56$\pm$0.51} & 10.88$\pm$0.10 & 75.28$\pm$0.55 \\
 & LC-CBM (ours) & \textbf{86.47$\pm$0.02} & 71.47$\pm$0.29 & \textbf{0.97$\pm$0.07} & \textbf{84.45$\pm$0.06} \\
\midrule
\multirow{2}{*}{CIFAR-100}
 & ECBM          & 84.67$\pm$0.04 & 37.56$\pm$0.83 & 12.09$\pm$0.03 & 33.30$\pm$0.86 \\
 & LC-CBM (ours) & \textbf{85.02$\pm$0.03} & \textbf{40.03$\pm$0.47} & \textbf{1.47$\pm$0.09} & \textbf{51.03$\pm$0.27} \\
\bottomrule
\end{tabular}
\end{table}

\newpage

\section{Experiments for concept leakage}
\label{sec:lf_leak}

A concept bottleneck leaks when its concept layer passes information about the target that the concept labels themselves do not contain
\citep{margeloiu2021concept,mahinpei2021promises}.
Since LC-CBM models the dependencies between concepts, it could in principle
leak more than models with independent concepts.
We test this with two measures of leakage:
Shuffled-concept accuracy and Own versus all-revealed accuracy.

\paragraph{Shuffled-concept accuracy.}
Let $\{(x_i, c_i, y_i)\}_{i=1}^{n}$ be the training set and $\pi$ a random permutation of $\{1,\dots,n\}$. We retrain each
model with the same hyperparameters on $\{(x_i, c_{\pi(i)}, y_i)\}$, so that the concept labels keep their marginal
and joint statistics but carry no information about $x$ or $y$ (the validation set is permuted in the same way).
$\mathrm{Acc}_{\mathrm{shuf}}$ is the test accuracy of the retrained model.
A model whose predictions go through the meaning of its concepts should be at chance (0.10 on CIFAR-10, 0.01 on
CIFAR-100); a higher value is leakage \citep{mahinpei2021promises}. 
For LF-CBM we permute the CLIP
similarities and disable the interpretability cut-off, which would otherwise remove every concept.

\paragraph{Own versus all-revealed accuracy.}
For the models trained on the true labels, with $c^\ast_i$ the oracle concepts of test image $i$, we compare
\[
\mathrm{Acc}_{\mathrm{own}} = \frac{1}{N}\sum_{i=1}^{N}\mathbb{I}\Big[\arg\max_y\, p(y\mid x_i) = y_i\Big],
\qquad
\mathrm{Acc}_{\mathrm{all}} = \frac{1}{N}\sum_{i=1}^{N}\mathbb{I}\Big[\arg\max_y\, p(y\mid c^\ast_i) = y_i\Big],
\]
where $p(y\mid x)=\mathbb{E}_{c\sim p(c\mid x)}[p(y\mid c)]$ uses the model's own concept predictions and
$p(y\mid c^\ast)$ is the head applied to the oracle concepts. For LF-CBM each concept activation is set to the mean
activation of training images with the same label. If the head uses what the concepts mean,
$\mathrm{Acc}_{\mathrm{all}}\ge\mathrm{Acc}_{\mathrm{own}}$. If it relies on information outside the concept values,
$\mathrm{Acc}_{\mathrm{all}}$ drops.

\paragraph{Results.}
Table~\ref{tab:lf_leak} shows the results.
With shuffled concept labels, LC-CBM and Hard CBM stay at or near chance, and SCBM leaks slightly on CIFAR-100 (0.050), whereas LF-CBM remains almost as accurate as with the true labels.
Revealing all concepts raises the accuracy of LC-CBM, Hard CBM and SCBM, but sharply lowers that of LF-CBM.
LC-CBM shows no sign of leakage in these tests, and modeling the dependencies between
concepts does not create a leakage channel.
LF-CBM, in contrast, leaks strongly, because its head reads class information
from continuous concept activations.

\begin{table*}[h]
\centering
\small
\setlength{\tabcolsep}{4pt}
\caption{Concept leakage with complete concept labels (mean $\pm$ std over 3 seeds).
$\mathrm{Acc}_{\mathrm{shuf}}$: accuracy after training on concept labels permuted across images (chance: 0.10 on
CIFAR-10, 0.01 on CIFAR-100). $\mathrm{Acc}_{\mathrm{own}}$: accuracy with the model's own concept predictions.
$\mathrm{Acc}_{\mathrm{all}}$: accuracy with all concepts set to their oracle values. SCBM uses the amortized
covariance on CIFAR-10 and the global covariance on CIFAR-100. Bold: lowest $\mathrm{Acc}_{\mathrm{shuf}}$ and highest
$\mathrm{Acc}_{\mathrm{all}}$.}
\label{tab:lf_leak}
\begin{tabular}{lcccccc}
\toprule
 & \multicolumn{3}{c}{CIFAR-10} & \multicolumn{3}{c}{CIFAR-100} \\
\cmidrule(lr){2-4}\cmidrule(lr){5-7}
Model & $\mathrm{Acc}_{\mathrm{shuf}}\downarrow$ & $\mathrm{Acc}_{\mathrm{own}}$ & $\mathrm{Acc}_{\mathrm{all}}\uparrow$
      & $\mathrm{Acc}_{\mathrm{shuf}}\downarrow$ & $\mathrm{Acc}_{\mathrm{own}}$ & $\mathrm{Acc}_{\mathrm{all}}\uparrow$ \\
\midrule
LC-CBM (ours) & \textbf{0.113{\scriptsize$\pm$0.011}} & 0.877{\scriptsize$\pm$0.001} & \textbf{0.894{\scriptsize$\pm$0.001}} & \textbf{0.010{\scriptsize$\pm$0.000}} & 0.655{\scriptsize$\pm$0.002} & 0.742{\scriptsize$\pm$0.001} \\
Hard CBM & 0.129{\scriptsize$\pm$0.029} & 0.867{\scriptsize$\pm$0.002} & 0.893{\scriptsize$\pm$0.002} & \textbf{0.010{\scriptsize$\pm$0.000}} & 0.620{\scriptsize$\pm$0.003} & \textbf{0.751{\scriptsize$\pm$0.002}} \\
Amortized SCBM or Global SCBM& 0.125{\scriptsize$\pm$0.009} & 0.855{\scriptsize$\pm$0.012} & 0.893{\scriptsize$\pm$0.004} & 0.050{\scriptsize$\pm$0.019} & 0.643{\scriptsize$\pm$0.003} & \textbf{0.751{\scriptsize$\pm$0.002}} \\
LF-CBM & 0.798{\scriptsize$\pm$0.007} & 0.867{\scriptsize$\pm$0.000} & 0.247{\scriptsize$\pm$0.003} & 0.653{\scriptsize$\pm$0.002} & 0.648{\scriptsize$\pm$0.001} & 0.075{\scriptsize$\pm$0.001} \\
\bottomrule
\end{tabular}
\end{table*}

\newpage

% =============================================================================================

\section{Further examples of interpretation}
\label{app:interpretation}

\begin{figure}[h]
  \centering
  \includegraphics[width=0.8\linewidth]{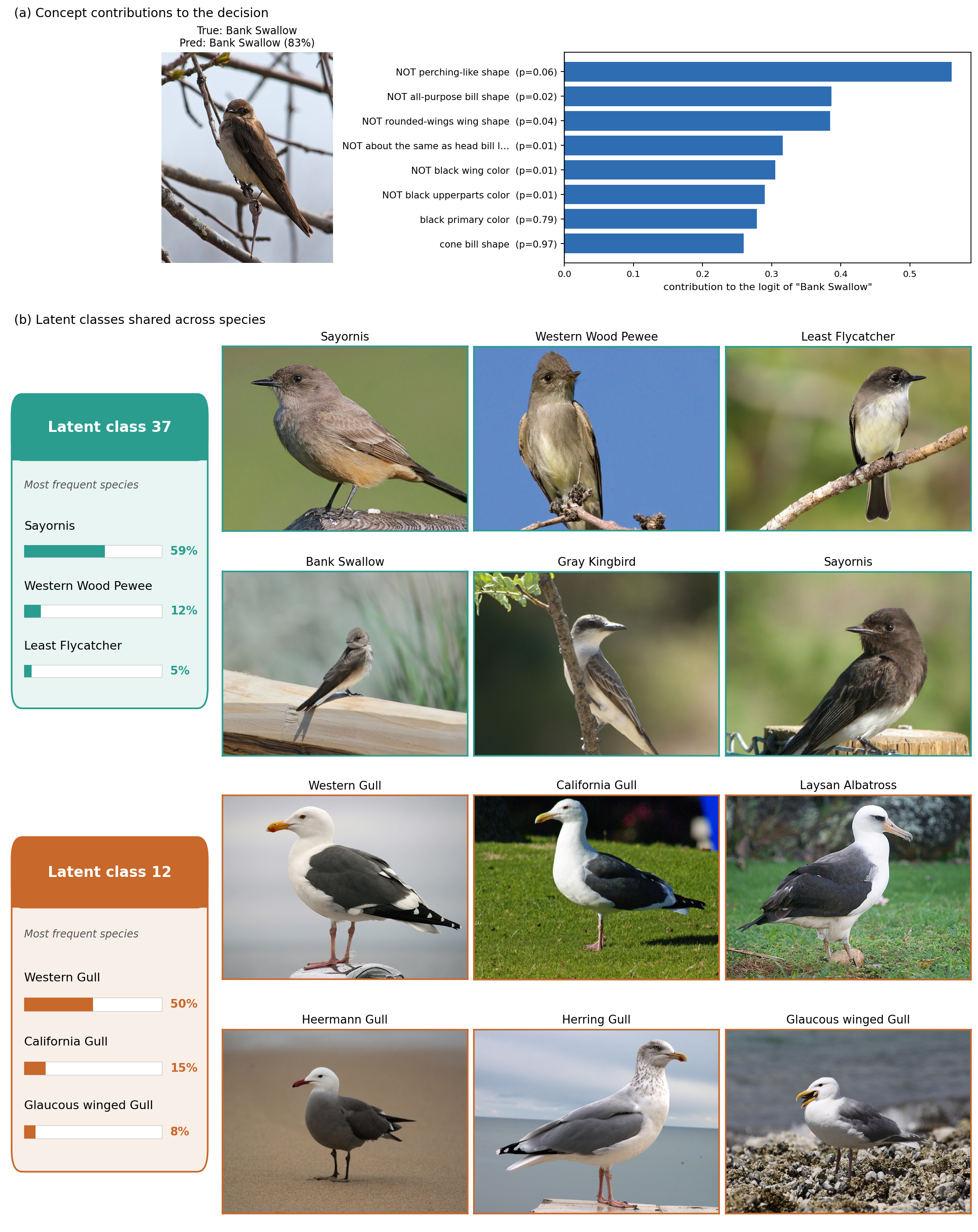}
  \caption{Further examples of interpreting LC-CBM on CUB, as in Figure~\ref{fig:interp}.
  (a)~The eight largest concept contributions $W[y,j]\,(p(c_j{=}1\mid x) - \bar p_j)$ to the logit of the predicted species, a Bank Swallow.
  Parentheses give $p(c_j{=}1\mid x)$; ``NOT'' marks concepts less likely than on average.
  (b)~Test images of different species whose most probable latent class is class~37 (top) and class~12 (bottom).
  The cards list the three most frequent species in each class over the train and test images.}
  \label{fig:interp_app}
\end{figure}

Figure~\ref{fig:interp_app} gives further examples of the analysis in Section~\ref{sec:interpretation}.
We use the same LC-CBM trained on CUB (seed 42) and its test images.

\paragraph{Explaining a decision.}
Figure~\ref{fig:interp_app}(a) explains a correct prediction for a Bank Swallow.
In contrast to the Rose-breasted Grosbeak in Figure~\ref{fig:interp}(a), the decision rests almost entirely on absent concepts.
The bird does not have a perching-like shape, an all-purpose bill, or rounded wings, and it lacks black wings and black upperparts.
These features separate swallows from the many perching birds in CUB, which are similar in size and color but differ in shape.

\paragraph{Latent classes.}
Figure~\ref{fig:interp_app}(b) shows two further latent classes.
Latent class~37 joins phoebes, pewees, flycatchers, and a kingbird with the Bank Swallow, all small, plain, brownish-gray birds.
Latent class~12 joins several gulls with the Laysan Albatross, which shares their white body and dark back.
As in Figure~\ref{fig:interp}(b), each class groups different species that look alike in terms of the annotated concepts.
Latent class~37 contains only five species with at least three images in the class, so its last panel shows a second phoebe.

\end{document}